\documentclass[sigconf]{acmart}
\usepackage{amsthm}
\usepackage{amsmath}
\usepackage{graphicx}
\usepackage{multirow}
\usepackage{tabularx}
\usepackage{subfigure}
\usepackage{pifont}
\usepackage{color}

\AtBeginDocument{%
  }

\setcopyright{acmlicensed}
\copyrightyear{2018}
\acmYear{2018}
\acmDOI{XXXXXXX.XXXXXXX}
\acmConference[Conference acronym 'XX]{Make sure to enter the correct
  conference title from your rights confirmation emai}{June 03--05,
  2018}{Woodstock, NY}
\acmISBN{978-1-4503-XXXX-X/18/06}

\begin{document}

\title{FUGNN: Harmonizing Fairness and Utility in Graph Neural Networks}
\author{Renqiang Luo, Huafei Huang, Shuo Yu*, Zhuoyang Han}
\affiliation{
  \institution{Dalian University of Technology, China}
  \country{}
}
\email{{lrenqiang, hhuafei}@outlook.com}
\email{shuo.yu@ieee.org,42217022@mail.dlut.edu.cn}

\author{Estrid He, Xiuzhen Zhang}
\author{Feng Xia}
\affiliation{%
    \institution{RMIT University, Austalia}
    \country{}
}
\email{{estrid.he, xiuzhen.zhang}@rmit.edu.au}
\email{f.xia@ieee.org}

\thanks{This paper is accepted by KDD 2024.}
\thanks{* Corresponding Author}
\renewcommand{\shortauthors}{Luo, et al.}

\begin{abstract}
  Fairness-aware Graph Neural Networks (GNNs) often face a challenging trade-off, where prioritizing fairness may require compromising utility. 
  In this work, we re-examine fairness through the lens of spectral graph theory, aiming to reconcile fairness and utility within the framework of spectral graph learning. 
  We explore the correlation between sensitive features and spectrum in GNNs, using theoretical analysis to delineate the similarity between original sensitive features and those after convolution under different spectra. 
  Our analysis reveals a reduction in the impact of similarity when the eigenvectors associated with the largest magnitude eigenvalue exhibit directional similarity. 
  Based on these theoretical insights, we propose FUGNN, a novel spectral graph learning approach that harmonizes the conflict between fairness and utility. 
  FUGNN ensures algorithmic fairness and utility by truncating the spectrum and optimizing eigenvector distribution during the encoding process. 
  The fairness-aware eigenvector selection reduces the impact of convolution on sensitive features while concurrently minimizing the sacrifice of utility. 
  FUGNN further optimizes the distribution of eigenvectors through a transformer architecture. 
  By incorporating the optimized spectrum into the graph convolution network, FUGNN effectively learns node representations. 
  Experiments on six real-world datasets demonstrate the superiority of FUGNN over baseline methods.
  The codes are available at https://github.com/yushuowiki/FUGNN.
\end{abstract}



\maketitle

\section{Introduction}
With the prevalence of graph-structured data in real-world applications, graph neural networks (GNNs) have shown promising performance in high-stake domains \cite{xia2024coupled, tu2023deep}, such as loan approval \cite{cheng2023regulating}, disaster response \cite{xia2023cengcn}, criminal justice \cite{feng2023criminal}, and medical diagnoses \cite{albahri2023a}. 
In these applications, certain features (e.g., gender, race, age, and region) are legally protected to prevent abuse and are regarded as sensitive features \cite{garg2022handling}.
However, GNNs may produce biased predictions that discriminate against particular subgroups characterized by these sensitive features \cite{luo2024fairgt}.
For example, GNNs may cause racial discrimination in underdiagnoses \cite{chen2023algorithmic} or gender discrimination in low-interest loans \cite{kallus2022assessing}.
Hence, mitigating discrimination induced by GNNs to achieve fairness remains as a critical challenge in this domain.
\begin{figure}[H]
	\vspace{-0.5em}
  \centering
	\includegraphics[width=0.3\textwidth]{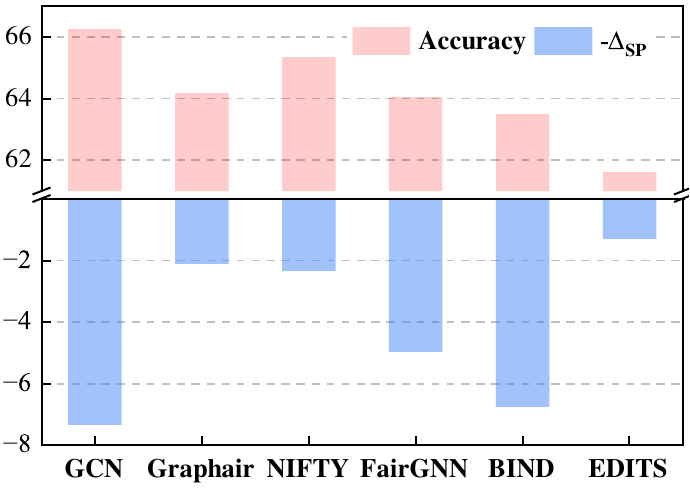}
    \caption{The sacrificed utility of fairness-aware GNNs. The utility is measured by the prediction accuracy and the fairness is measured by the -$\Delta_\text{SP}$.}
  \vspace{-0.5em}
    \label{fig:background_1}
  \vspace{-0.5em}
\end{figure}

Various efforts have been devoted to developing fairness-aware GNNs, aiming to control the degree to which a model depends on sensitive features, measured by independence criteria such as statistical parity and equality opportunity \cite{pessach2022a}.
Different controlling techniques have been proposed \cite{kang2022rawlsgcn}, including weighting perturbation \cite{jiang2023chasing}, embedding adjustment \cite{khajehnejad2022crosswalk}, pre-processing dataset \cite{abdelrazek2023fairup}, and loss function regularization \cite{dong2022edits}. 
These methods aim to promote the fairness but often come with a trade-off in utility, generally measured by predicting accuracy \cite{dong2023fairness}. 
Figure \ref{fig:background_1} compares the vanilla graph convolution network, a widely adopted graph neural architecture, and five fairness-aware GNNs in terms of utility (accuracy) and fairness (independence on sensitive features, i.e., -$\Delta_\text{SP}$) on a real-world dataset (i.e., Pokec-z). 
Although these fairness-aware GNNs reduce the dependence on sensitive features, utility is generally compromised. 

In this paper, we study fairness in GNNs from a new perspective, i.e., spectral theory, inspired by recent work on the expressive spectral filters and spectrum analysis of the underlying graph matrix (e.g., adjacency or Laplacian matrix and their normalized forms) \cite{bastos2023learnable}.
We ask the question: \textit{is it possible to harmonize the conflict between utility and fairness through graph spectral analysis?}
The key to a fairness-aware spectral filter lies in revealing the relationship between the fairness of a model and the graph spectrum. 
Therefore, our first step is to perform theoretical analysis to quantify the extent of a model's protection over sensitive features, via comparing the similarity between the original representations of sensitive features at the input layer and their deep representations after convolution using spectral theory. 

Our theoretical analysis reveal two findings:
(1) The similarity between original sensitive features and their deep representations after convolution is best captured by the eigenvector corresponding to the largest eigenvalue.
This motivates us to select principal eigenvalues as the spectrum for convolution, through which fairness in prediction results can be maintained.
This also ensures that model utility can be preserved since spectral graph theory indicates that the largest non-zero eigenvalues are linked to the geometry of the graph (including algebraic connectivity and spectral radius) \cite{kreuzer2021rethinking};
(2) the impact of non-principal eigenvectors on the fairness of a model diminishes exponentially with an increase in the number of convolutional layers. 
This motivates us to remove these components to guide the model to focus on those principal eigenvectors. 

Based on the theoretical analysis, we present a novel approach, \textbf{FUGNN} (harmonizing \textbf{F}airness and \textbf{U}tility \textbf{GNN}), which promotes fairness in graph learning at minimal cost to utility via spectrum modification. 
Our spectrum modification strategy is designed based on our two findings above, that is, harmonizing utility and fairness through selecting the spectral component that has the highest impact on utility and fairness. 
Specifically, FUGNN first computes $K$ largest magnitude eigenvalues along with corresponding eigenvectors from the adjacency matrix.
This subset of eigenvalues expresses convolution corresponding to sensitive features while mitigating the influence of non-principle eigenvectors.
Then, FUGNN employs a transformer architecture to optimize the distribution of eigenvectors, ensuring the independence of sensitive features during convolution.
Finally, our method incorporates the optimized spectrum in graph convolution, and obtains fair node representations. 
In summary, our contributions are outlined below:
\vspace{-2em}

\begin{itemize}
    \item \textbf{A synergistic approach to fairness and utility.} 
    We present FUGNN, a spectral graph learning method that harmonizes the conflict between fairness and utility in fairness-aware GNNs. 
    FUGNN strategically mitigates the convolution impact on sensitive features while achieving high utility.
    \item \textbf{Theoretical insights into sensitive feature expression.} 
    We study the model fairness preservation from a spectral perspective through rigorous theoretical analysis. 
    Our findings provide insights about the correlation between the graph spectrum and the deep representations of sensitive features within a graph model, forming the foundation of the proposed FUGNN.
    \item \textbf{A novel eigenvalue selection mechanism for preserving model fairness.} 
    Building on our theoretical findings, we introduce an innovative eigenvalue selection mechanism that can truncate the spectrum to ensure fairness without compromising utility, providing a nuanced approach to mitigating fairness challenges within GNNs. 
    \item \textbf{Empirical validation through extensive evaluations.} 
    To verify the effectiveness of the proposed FUGNN approach, we conduct comprehensive empirical evaluations on six real-world datasets. 
    The results demonstrate the superior performance of FUGNN in achieving both fairness and utility when compared to state-of-the-art fairness-aware GNNs.
\end{itemize}

\section{Related Work}

\subsection{Spectral GNNs}
Spectral GNNs process features through filters, including well-known models like GCN \cite{kipf2017semi}, SGC \cite{wu2019simplifying}, and S$^2$GC \cite{zhu2021simple}, which rely on eigenvectors of the (normalized) Laplacian for Graph Fourier Transform.
Polynomials are employed to approximate the spectral graph convolution, and are optimized to enhance the utility of GNNs.
JacobiConv \cite{wang2022how} uses the concept of an orthogonal basis, aligning its weight function with the graph signal density in the spectrum, and improves GNN utility through a novel polynomial coefficient decomposition technique.
ChebNetII \cite{he2022convolutional} is a novel GNN model that utilizes Chebyshev interpolation, focusing on the Chebyshev polynomials to achieve superior utility. 
In terms of node features, it is incorporated into several spectral graph learning.
CSBM-G \cite{wei2022understanding} imposes a Gaussian assumption on node features to better capture nonlinear relations in graph-structured data, which is significant when node features are more information than the graph structure.
FE-GNN \cite{sun2023feature} conducts a comprehensive examination of the dominant feature space in representation learning based on spectral model, optimizing GNN utility through the perspective of the dominant feature space. 
In conventional GNN architectures, the interdependence between different channels of the filter can impede the utility of more potent filters.
To address this, PDF \cite{yang2023towards} introduces a novel normalized adjacency matrix utility capable of leveraging an expanded set of bases and learnable filter coefficients, thereby enhancing responsiveness to the filter. 
Specformer \cite{bo2023specformer} further extends this methodology by broadening its applicability to point-level tasks.
However, these spectral filters do not explicitly consider algorithmic fairness. 
They disregard the interdependence between sensitive features and other features, potentially introducing bias into the filter.

\subsection{Fairness-aware GNNs}
The fairness of GNNs has gained substantial traction, primarily focusing on examination of discrimination associated with specific sensitive features. 
Predominant fairness-aware GNNs revolve around protecting the independence of sensitive features using pre-processing and in-processing methods \cite{caton2020fairness}.
Some previous methods improve fairness by pre-processing the dataset \cite{ling2023learning, guo2023fair}. 
For instance, Graphair \cite{ling2023learning} autonomously identifies fairness-aware augmentations from input graph data, aiming to circumventing sensitive features while preserving the integrity of other valuable data.
FairAC \cite{guo2023fair} shows a fair feature completion approach to address information gaps and acquire equitable node embeddings for graphs lacking features.
As an integral component of GNNs aimed at converting networks into low-dimensional vectors, fair embeddings as pre-processing methods play a crucial role in protecting sensitive features \cite{hou2020network}.
DeBayes \cite{buyl2020debayes} decreases bias in the embeddings by introducing biased information into the prior of the conditional network embedding.
To mitigate bias in node embeddings, FairSample \cite{cong2023fairsample} enhances fairness through a regularization objective.
Some in-processing methods use loss functions to constrain the influence of algorithms on sensitive features \cite{dong2022edits, dai2023learning}. 
EDITS \cite{dong2022edits} devises a loss function targeting the node output aimed to mitigate biases within the input feature network, fostering fairer GNN outcomes. 
FairGNN \cite{dai2023learning} is designed to limit the influence of sensitive features using an estimation function and adversarial debiasing loss function.
However, these algorithms often enhance fairness at the expense of the utility of algorithms.

\section{Preliminaries}
\subsection{Notations}
Unless otherwise specified, we denote sets with copperplate uppercase letters (i.e., $\mathcal{A}$), matrices with bold uppercase letters (i.e., $\mathbf{A}$), and vectors with bold lowercase letters (i.e., $\mathbf{a}$).

We denote an undirected graph as $\mathcal{G} = (\mathcal{V}, \mathbf{X})$, where $\mathcal{V}$ is the set of $n$ nodes in the graph, $\mathbf{X} \in \mathbb{R}^{n \times d}$ is the node feature matrix, and $d$ is the feature dimension.
For the $l$-th graph convolution layer, denoting its output node utility as $\mathbf{H}^{(l)}$, we generalize spectral graph convolution as follows \cite{klicpera2019predict,chen2020simple}:
\begin{equation}
    \mathbf{H}^{(l)} = (1-\theta)\mathbf{S} \mathbf{H}^{(l-1)} + \theta \mathbf{H}^{(0)},
\label{equ:spectral_graph}
\end{equation}
where $\theta \in (0,1)$, $\mathbf{H}^{(0)} = f_\Theta(\mathbf{X})$, and $\mathbf{S}$ is the adjacency matrix.
$h \in \mathbb{R}^{1 \times n}$ is one channel in the filter that corresponds to one dimension of $\mathbf{H^{(0)}}$.

In particular, we denote the sensitive feature channel as $h_{sen}$.
Based on the generalized formulation in Equation (1), we conduct fairness analysis on existing graph convolutions from the perspective of the graph's spectrum.
Assume $\mathbf{S} \in \mathbb{R}^{n \times n}$ is a symmetric matrix with real-valued entries.
$|\lambda_1| \geq |\lambda_2| \geq ......  \geq |\lambda_n|$ are $n$ real eigenvalues, and $\mathbf{p}_i$ ($i \in \{1, 2, ......, n\}$) are the corresponding eigenvectors.
After one layer of convolution, $h_{sen}$ is represented as $\mathbf{S}h_{sen}$. After $l$ layers convolution, it is represented as $\mathbf{S}^l h_{sen}$.
The cosine similarity between $\mathbf{S}^l h_{sen}$ and $h_{sen}$, denoted as $cos(\langle\mathbf{S}^l h_{sen}, h_{sen}\rangle)$, reflects the similarity between original sensitive features and the sensitive features after $l$ layers of convolution.
A higher cosine similarity indicates stronger protection of sensitive feature independence, reflecting higher fairness in GNNs.
Furthermore, we measure the eigenvector's influence on the sensitive features by $\sum_{i=1}^{K}cos(\langle h_{sen},\mathbf{p}_i\rangle)$, where $K$ is the number of selection eigenvectors.

\subsection{Fairness Evaluation Metrics}
In this subsection, we present two definitions of fairness for the binary label $y \in \{0,1\}$ and sensitive features $s \in \{0,1\}$.
We use $\hat{y} \in \{0,1\}$ to represent the predicted class label.

\textit{Definition 1. Statistical Parity (i.e., Demographic Parity, Independence) }\cite{dwork2012fairness}. 
Statistical parity requires the predictions to be independent of the sensitive features $s$. 
It can be formally written as:
\begin{equation}
    \mathbb{P}(\hat{y}|s=0)=\mathbb{P}(\hat{y}|s=1).
\label{equ:SP}
\end{equation}

When both the predicted labels and sensitive features are binary, the extent of statistical parity can be quantified by $\Delta_\text{SP}$, defined as follows:
\begin{equation}
    \Delta_{SP}=|\mathbb{P}(\hat{y}=1|s=0)-\mathbb{P}(\hat{y}=1|s=1)|.
\label{equ:delta_SP}
\end{equation}

The $\Delta_\text{SP}$ measures the acceptance rate difference between the two sensitive subgroups.

\textit{Definition 2. Equal Opportunity} \cite{hardt2016equality}. 
Equal opportunity necessitates that the likelihood of an instance belonging to a positive class leading to a positive outcome should be equitable for all members within subgroups. 
For individuals with positive ground truth labels, it is necessary for positive predictions to be devoid of any dependence on sensitive features.
This principle can be mathematically expressed as follows:
\begin{equation}
    \mathbb{P}(\hat{y}=1|y=1,s=0)=\mathbb{P}(\hat{y}=1|y=1,s=1).
\label{equ:EO}
\end{equation}

Fairness-aware GNNs prevent the allocation of unfavorable predictions to individuals who are eligible for advantageous ones solely based on their sensitive subgroup affiliation.
In particular, $\Delta_\text{EO}$ quantifies the extent of deviation in predictions from the ideal scenario where equality of opportunity is satisfied.
To quantitatively assess euqal opportunity, we employ the following metric:
\begin{equation}
    \Delta_{EO}=|\mathbb{P}(\hat{y}=1|y=1,s=0)-\mathbb{P}(\hat{y}=1|y=1,s=1)|.
\label{equ:delta_EO}
\end{equation}

Both probabilities are evaluated on the test set.

\section{FUGNN: Theoretical Discovery}
\label{sec:TD}
In this section, we present our theoretical findings that underpin FUGNN.  
We analyze the correlation between sensitive features and the graph spectrum. 
The independence of a model on sensitive features can be reflected by the cosine similarity between $\mathbf{S}^l h_{sen}$ and $h_{sen}$, where a higher cosine similarity indicates stronger protection of sensitive feature independence. 
Hence, we analyze the relationship between the graph spectrum and the term $cos(\langle\mathbf{S}^l h_{sen}, h_{sen}\rangle)$. 
We aim to identify the components from the entire graph spectrum that have the most significant impact on the fairness of a model. 
We consider three different spectral components: 
(1) a single eigenvector with the largest magnitude of eigenvalue; 
(2) multiples eigenvectors with the (same) largest magnitude of eigenvalue; 
and (3) non-principal eigenvectors (i.e., eigenvectors with small eigenvalues). 

\subsection{Single Eigenvector Corresponding to the Largest Magnitude Eigenvalue}

\textit{Lemma 1}.
Assume $\mathbf{S} \in \mathbb{R}^{n \times n}$ is a symmetric matrix with real-valued entries.
The eigenvalue are ordered as $|\lambda_1| > |\lambda_2| \geq ......  \geq |\lambda_n|$, and $\mathbf{p}_i$ ($i \in \{1, 2, ......, n\}$) are corresponding eigenvectors.
Then the following equation holds:
\begin{equation}
    \lim_{l \to \infty} cos(\langle\mathbf{S}^l h_{sen}, h_{sen}\rangle) = cos(\langle h_{sen}, \mathbf{p}_1\rangle).
\label{equ:fairness_definition_single}
\end{equation}

\textit{Proof}. 
Since $\mathbf{S} \in \mathbb{R}^{n \times n}$ is a symmetric matrix, the eigendecomposition of $\mathbf{S}$ can be written as $\mathbf{S} = \mathbf{P} \varLambda \mathbf{P}^\top$ with $\mathbf{P} = (\mathbf{p}_1, \mathbf{p}_2,......,\mathbf{p}_n)$, $\Vert \mathbf{p}_i \Vert = 1$ ($i \in \{1, 2, ......, n\}$) and $\varLambda = diag(\lambda_1, \lambda_2, ......, \lambda_n)$.
The cosine similarity can be expressed as:
\begin{equation}
    \nonumber
    \begin{aligned}
	    cos(\langle h_{sen}, \mathbf{p}_1\rangle)
      &=\frac{h_{sen}^\top \mathbf{p}_1}{\Vert h_{sen} \Vert \Vert \mathbf{p}_1 \Vert}=\frac{h_{sen}^\top \mathbf{p}_1}{\Vert h_{sen} \Vert}\\
	    &=\frac{h_{sen}^\top \mathbf{p}_1}{\sqrt{h_{sen}^\top h_{sen}}}=\frac{h_{sen}^\top \mathbf{p}_1}{\sqrt{(\mathbf{P}^\top h_{sen})^\top \mathbf{P}^\top h_{sen}}}\\
      &=\frac{h_{sen}^\top \mathbf{p}_1}{\sqrt{\sum_{i=1}^n(h_{sen}^\top \mathbf{p}_i)^2}}.
    \end{aligned}
\end{equation}

Assuming $\alpha_i = h_{sen}^\top \mathbf{p}_i$, the weight of $h_{sen}$ on $\mathbf{p}_i$ when representing $h_{sen}$ with the set of orthonormal bases $\mathbf{p}_i$ ($i \in \{1, 2, ......, n\}$), then:
\begin{equation}
    \nonumber
    cos(\langle h_{sen}, \mathbf{p}_1\rangle)=\frac{\alpha_1}{\sqrt{\sum^n_{i=1}\alpha_i^2}}.
\end{equation}

When $l \rightarrow \infty$, we have:
\begin{equation}
    \nonumber
    \begin{aligned}
	&{\lim_{l \rightarrow \infty}}cos(\langle\mathbf{S}^l h_{sen}, h_{sen}\rangle)\\    
    =&\lim_{l\rightarrow\infty}\frac{(\mathbf{S}^l h_{sen})^\top h_{sen}}{\Vert \mathbf{S}^l h_{sen} \Vert \Vert h_{sen} \Vert}\\
	=&\lim_{l\rightarrow\infty}\frac{(\mathbf{S}^l h_{sen})^\top h_{sen}}{\sqrt{(\mathbf{S}^l h_{sen})^\top \mathbf{S}^l h_{sen}} \sqrt{h_{sen}^\top h_{sen}}}\\
    =&\lim_{l\rightarrow\infty}\frac{(\mathbf{P} \varLambda^l \mathbf{P}^\top h_{sen})^\top h_{sen}}{\sqrt{(\mathbf{P} \Lambda^l \mathbf{P}^\top h_{sen})^\top (\mathbf{P} \varLambda^l \mathbf{P}^\top h_{sen})} \sqrt{h_{sen}^\top h_{sen}}}\\
    =&\lim_{l\rightarrow\infty}\frac{(\mathbf{P}^\top h_{sen})^\top \varLambda^l (\mathbf{P}^\top h_{sen})}{\sqrt{(\mathbf{P}^\top h_{sen})^\top \varLambda^{2l} (\mathbf{P}^\top h_{sen})} \sqrt{h_{sen}^\top h_{sen}}}\\
    =&\lim_{l\rightarrow\infty}\frac{\sum^n_{i=1}\alpha_i^2 \lambda_i^l}{\sqrt{\sum^n_{i=1}\alpha_i^2 \lambda_i^{2l}}\sqrt{\sum^n_{i=1}\alpha_i^2}}\\
	=&\lim_{l\rightarrow\infty}\frac{\alpha_1^2 + \sum^n_{i=2}\alpha_i^2 (\frac{\lambda_i}{\lambda_1})^l}{\sqrt{\alpha_1^2 + \sum^n_{i=2}\alpha_i^2 (\frac{\lambda_i}{\lambda_1})^{2l}}\sqrt{\sum^n_{i=1}\alpha_i^2}}\\
    =&\frac{\alpha_1}{\sqrt{\sum^n_{i=1}\alpha_i^2}}.
    \end{aligned}
\end{equation}
\rightline{$\square$}

The above theorem shows that the term $cos(\langle\mathbf{S}^l h_{sen}, h_{sen}\rangle)$ is well captured by the spectral component $\mathbf{p}_1$. 
This indicates that the principal component carries the most information about the fairness of a model. 

\subsection{Multiple Eigenvectors Corresponding to the Largest Magnitude Eigenvalue}
The matrix $\mathbf{S}$ can possess multiple eigenvectors associated with the largest magnitude eigenvalue, and thus, \textit{Lemma 1} does not hold. Next, we investigate such cases. 

\textit{Lemma 2}.
Assume that there are multiple eigenvectors corresponding to the largest magnitude eigenvalue in the real-valued entries.
The eigenvalues are ordered as $|\lambda_1| = |\lambda_2| = ...... = |\lambda_j| > |\lambda_{j+1}| \geq ...... \geq |\lambda_n|$, and $\mathbf{p}_i \in \mathbb{R}^n, i \in \{1, 2, ......, n\}$ are corresponding eigenvectors.
Then the following equation holds:
\begin{equation}
    \lim_{l \to \infty} cos(\langle\mathbf{S}^l h_{sen}, h_{sen}\rangle) \geq \frac{1}{\sqrt{j}}\sum_{i=1}^j{cos(\langle h_{sen}, \mathbf{p}_i\rangle)}.
\label{equ:fairness_definition_multi}
\end{equation}

\textit{Proof}. 
\begin{equation}
    \nonumber
    \begin{aligned}
	&{\lim_{l \rightarrow \infty}}cos(\langle\mathbf{S}^l h_{sen}, h_{sen}\rangle)\\
	=&\lim_{l\rightarrow\infty}\frac{\sum^n_{i=1}\alpha_i^2 \lambda_i^l}{\sqrt{\sum^n_{i=1}\alpha_i^2 \lambda_i^{2l}}\sqrt{\sum^n_{i=1}\alpha_i^2}}\\
	=&\lim_{l\rightarrow\infty}\frac{\sum^j_{i=1}\alpha_i^2 + \sum^n_{i=j+1}\alpha_i^2 (\frac{\lambda_i}{\lambda_1})^l}{\sqrt{\sum^j_{i=1}\alpha_i^2 + \sum^n_{i=j+1}\alpha_i^2 (\frac{\lambda_i}{\lambda_1})^{2l}}\sqrt{\sum^n_{i=1}\alpha_i^2}}\\
    =&\frac{\sqrt{\sum_{i=1}^j\alpha_i^2}}{\sqrt{\sum^n_{i=1}\alpha_i^2}}.
    \end{aligned}
\end{equation}
\begin{figure*}[htbp]
	\centering
	\includegraphics[width=0.8\textwidth]{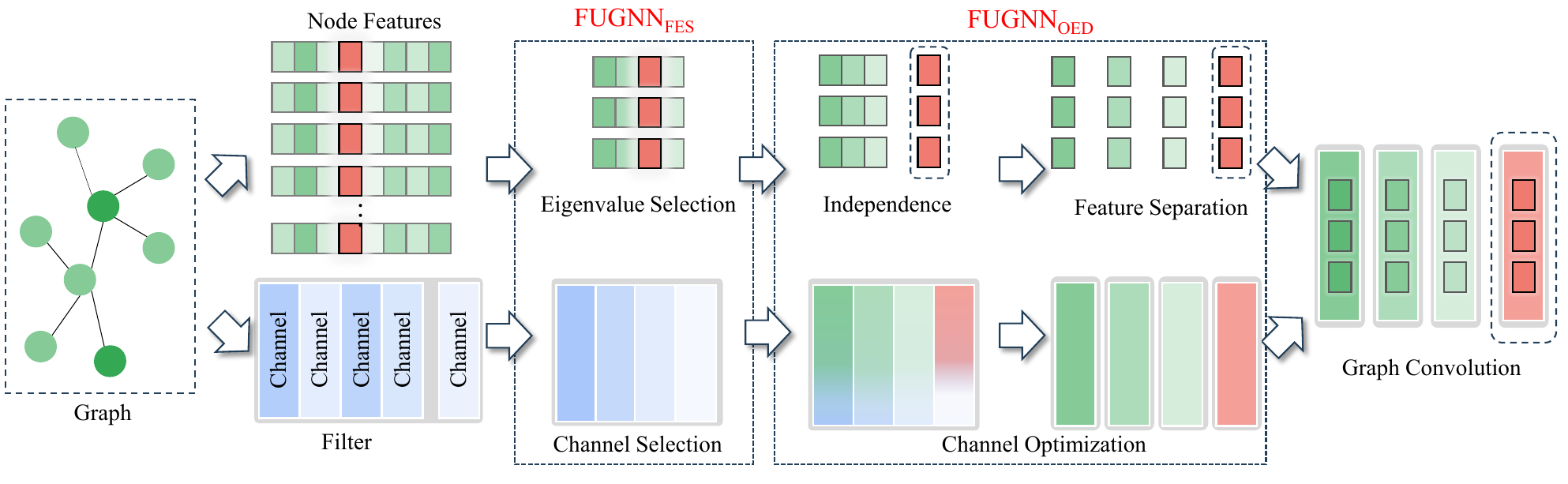}
    \caption{The framework of FUGNN.
    The model applies spectral truncation via eigenvalue selection and eigenvector distribution optimization with fairness considerations. 
    The top and bottom illustrate the two stages from the perspective of feature-level representations and feature channels, respectively.
    }
    \vspace{-0.5em}
    \label{fig:illustration}
    \vspace{-0.5em}
\end{figure*}

Then, considering $cos(\langle h_{sen}, \mathbf{p}_i\rangle)=\frac{\alpha_i}{\sqrt{\sum^n_{i=1}\alpha_i^2}}$ and the Cauchy-Schwarz Inequality, we have:
\begin{equation}
    \nonumber
    \begin{aligned}
	  &\lim_{l \to \infty} cos(\langle\mathbf{S}^l h_{sen}, h_{sen}\rangle) \\ 
    =& \frac{1}{\sqrt{j}}\frac{\sqrt{j*1^2\sum_{i=1}^j\alpha_i^2}}{\sqrt{\sum^n_{i=1}\alpha_i^2}} \\
    \geq &\frac{1}{\sqrt{j}}\frac{\sum_{i=1}^j\alpha_i}{{\sqrt{\sum_{i=1}^n\alpha_i^2}}} \\
    = &\frac{1}{\sqrt{j}}\sum \nolimits_{i=1}^j{cos(\langle h_{sen}, \mathbf{p}_i\rangle)}.
    \end{aligned}
\end{equation}

Especially, only if right-side terms $cos(\langle h_{sen}, \mathbf{p}_i\rangle)$ are equal, the equation holds.

\rightline{$\square$}

\textit{Lemma 2} shows that the term $cos(\langle\mathbf{S}^l h_{sen}, h_{sen}\rangle)$ is determined by multiple principal components.
The lower bound of the term $cos(\langle\mathbf{S}^l h_{sen}, h_{sen}\rangle)$ is a weighted sum of the multiple eigenvectors corresponding to the largest magnitude eigenvalue.
In particular, the similarity is maximum when the directions of these eigenvectors are similar.

\subsection{Non-principal Eigenvectors}
For an eigenvector of $\mathbf{S}$, if the corresponding eigenvalue $|\lambda_i| << |\lambda_1|$, the eigenvector is regarded as non-principal eigenvector.

\textit{Lemma 3}.
The influence of non-principal eigenvectors on sensitive features decays exponentially.

\textit{Proof}.
Because 
\begin{equation}
  \nonumber
  \begin{aligned}
  cos(\langle h_{sen}, \mathbf{p}_i\rangle)=\frac{\alpha_i}{\sqrt{\sum^n_{j=1}\alpha_j^2}},
  \end{aligned}
\end{equation}
and
\begin{equation}
  \nonumber
  \begin{aligned}
    &cos(\langle\mathbf{S}^l h_{sen}, h_{sen}\rangle)
    = &\frac{\alpha_1^2 + \sum^n_{i=2}\alpha_i^2 (\frac{\lambda_i}{\lambda_1})^l}{\sqrt{\alpha_1^2 + \sum^n_{i=2}\alpha_i^2 (\frac{\lambda_i}{\lambda_1})^{2l}}\sqrt{\sum^n_{i=1}\alpha_i^2}}.
  \end{aligned}
\end{equation}

Hence, the correlation between $\mathbf{p}_i$ and $cos(\langle\mathbf{S}^l h_{sen}, h_{sen}\rangle)$ is proportional to $(\frac{\lambda_i}{\lambda_1})^{l}$.
Because of $|\lambda_1| >>|\lambda_i|$, the correlation decays exponentially. 
It's even negligible with fewer layers.

\rightline{$\square$}

The above analysis proves that impact of non-principal eigenvectors on the deep representations of sensitive features (i.e., after convolution) diminishes exponentially.

In conclusion, sensitive features, after $l$ layers of convolution, are predominantly influenced by the eigenvector corresponding to the largest magnitude eigenvalue (or multiple eigenvectors corresponding to the largest magnitude eigenvalue).
Fortunately, as shown in previous studies, these principal eigenvectors are also most influential in the utility of a model \cite{kreuzer2021rethinking}. 
Thus, we argue that truncating the spectrum to include only principal eigenvectors can enhance the fairness of a model without compromising utility.
 
\section{FUGNN: Technical Details}
In this section, we introduce the design details of FUGNN.
Our approach involves three main components: fairness-aware eigenvalue selection, optimization of eigenvector distribution, and graph convolution.
Firstly, we compute the $K$ largest magnitude eigenvalues and their corresponding eigenvectors.
These principal eigenvectors are retained as they are shown to be effective in expressing the structure of the graph and in preserving sensitive features, as demonstrated by our theoretical analysis \cite{kreuzer2021rethinking}. 
Secondly, we undertake adjustments in the distribution of eigenvectors with the transformer architecture, which further harmonizes and improves the utility and fairness of the model.
Lastly, we assign the optimizing spectrum into graph convolution. 
Figure \ref{fig:illustration} illustrates the model architecture.

\subsection{Fairness-aware Eigenvalue Selection (FES)}
\label{sec:FES}
In Section~\ref{sec:TD}, we have shown that sensitive features, after $l$ layers convolution, primarily correlate with the largest magnitude eigenvalue and their corresponding eigenvectors. 
Hence, keeping these eigenvectors in the graph spectrum can promote the fairness of the model. 
Meanwhile, according to \textit{Lemma 2} and \textit{Lemma3}, the impact of a non-principal eigenvector on the model fairness is negligible after $l$ layer of convolution. We hypothesize that removing these non-principal eigenvectors can better guide the model to focus on the principal eigenvectors.

Thus, we propose to modify the graph spectrum by selecting those eigenvectors. 
We leverage the Arnoldi Package algorithm \cite{richard1998arpack} to obtain the eigenvalues of the adjacency matrix by computing a subset of $K$ eigenvalues and corresponding eigenvectors: 
\begin{equation}
  \begin{aligned}
  \mathbf{e}_{FES} = (\lambda_1, \lambda_2, ...... ,\lambda_K),\\
  \mathbf{P}_{FES} = (\mathbf{p_1}, \mathbf{p_2}, ...... , \mathbf{p_K}).
  \end{aligned}
\end{equation}
We choose Arnoldi Package due to its high computational efficiency and its accuracy as empirically demonstrated in \cite{cai2021a}.

Here, $K$ is a model hyperparameter representing the number of principal eigenvectors that are kept in the graph spectrum. 
In our empirical analysis in Section~\ref{sec:PA}, we show that the effectiveness of a model (i.e., fairness and utility) is influenced by the value $K$. 
At lower range of $K$ values (e.g., <10), a model maintains high effectiveness with only slight fluctuations. 
However, when too many non-principle eigenvectors are kept in the spectrum (e.g., $K>100$), the effectiveness of a model will decrease significantly. 

\subsection{Optimization of Eigenvectors Distribution (OED)}
By \textit{Lemma 2}, the maximum similarity between $h_{sen}$ and $\textbf{S}^lh_{sen}$ is intrinsically tied to the eigenvectors corresponding to the largest eigenvalue. 
That is, the model fairness is best preserved when the variation in the distribution of eigenvectors is minimal. 
Thus, to decrease the influence of convolutional layers on sensitive features, we optimize the eigenvalue distribution using Transformer.
Although the Transformer architecture has demonstrated effectiveness in achieving high model utility \cite{ying2021do}, the purpose of introducing this module to our proposed approach is different. 
We hope to achieve high model utility, but more importantly, to optimize the eigenvector distribution for the purpose of preserving fairness.

More specifically, we encode the eigenvalues $\lambda_{2i} \in \mathbf{e}_{FES}$ and $\lambda_{2i+1} \in \mathbf{e}_{FES}$  as follows:
\begin{equation}
    \begin{aligned}
        &\rho(\lambda_{2i}) = sin(\lambda_{2i}/10000^{2i/d}), \\
        &\rho(\lambda_{2i+1}) = cos(\lambda_{2i+1}/10000^{2i/d}).
    \end{aligned}
\label{equ:encoding_function}
\end{equation}

We denote $\mathbf{e}_{POS} = (\rho(\lambda_1), \rho(\lambda_2), ......, \rho(\lambda_K))$.
Then, the method uses multi-head self-attention (MHA) with layer normalization (LN).  
This modification facilitates the encoding of eigenvalues, enabling the capture of their inter-dependencies and generating valuable utility.
This eigenvalue encoding serves to alter the similarity between the eigenvector corresponding to the largest eigenvalue and other eigenvectors, thereby effectively enhancing the utility of sensitive features.
The formal characterization of the eigenvalue adjustment is provided as follows:
\begin{equation}
    \begin{aligned}
        &\mathbf{e}_{MHA}= \textbf{MHA}(\textbf{LN}(\mathbf{e}_{POS}))+\mathbf{e}_{POS}.
    \end{aligned}
\label{equ:MHA}
\end{equation}
\begin{equation}
  \begin{aligned}
      &\mathbf{e}_{OED}= \textbf{FFN}(\textbf{LN}(\mathbf{e}_{MHA}))+ \mathbf{e}_{MHA}.
  \end{aligned}
\label{equ:FFN}
\end{equation}

\subsection{Graph Convolution}
Finally, we assign optimizing spectrum based on learned basis $\mathbf{e}_{OED}$, and transform $\mathbf{H}^{(l-1)}$ into: 
\begin{equation}
  \mathbf{H'}^{(l-1)} = \mathbf{P}_{FES} \cdot (e_{OED} \odot \mathbf{P}_{FES}^\top \mathbf{H}^{(l-1)}).
\end{equation}

The graph convolution can be written as follows:
\begin{equation}
  \mathbf{H}^{(l)}= \sigma ((\mathbf{H}^{(l-1)} || \mathbf{H'}^{(l-1)}) \mathbf{W}^{(l-1)}),
\end{equation}
where $\mathbf{W}^{(l-1)}$ is the transformation, and $\sigma$ is the activation function.
By stacking multiple graph convolutional layers, FUGNN could learn node representation.

\section{Experiments}

\subsection{Datasets}
We showcase the effectiveness of our method on the downstream task, node classification, and adopt six real-world datasets for this task:
\begin{table*}
	\centering
	\caption{Comparison on accuracy and fairness ( $\Delta_\text{SP}$ and $\Delta_\text{EO}$ ) in percentage (\%) with six real world datasets. $\uparrow$ denotes the larger, the better; $\downarrow$ denotes the opposite. The best results are bold-faced.}
        \centering
        \tabcolsep=0.12cm
		\begin{tabular}{lccccccccc}
			\toprule
			\multirow{2}[2]{*}{\textbf{Methods}} & \multicolumn{3}{c}{\textbf{Income}} & \multicolumn{3}{c}{\textbf{Pokec-z}} & \multicolumn{3}{c}{\textbf{Pokec-n}} \\
            \cline{2-10}
			& ACC(\%) $\uparrow$  & $\Delta_\text{SP}$(\%) $\downarrow$   & $\Delta_\text{EO}$(\%) $\downarrow$  & ACC(\%) $\uparrow$  & $\Delta_\text{SP}$(\%) $\downarrow$   & $\Delta_\text{EO}$(\%) $\downarrow$ & ACC(\%) $\uparrow$  & $\Delta_\text{SP}$(\%) $\downarrow$ & $\Delta_\text{EO}$(\%) $\downarrow$\\
			\midrule
			\textbf{GCN}      & $74.73 \pm 2.54$  & $25.90 \pm 0.44$  & $32.30 \pm 2.78$  & $66.24 \pm 2.12$  & $7.32 \pm 1.48$  & $7.60 \pm 1.87$  & $66.53 \pm 2.84$   & $6.57 \pm 1.48$  & $5.33 \pm 0.42$  \\
      \textbf{GCNII}    & $76.24 \pm 2.46$  & $16.20 \pm 0.85$  & $25.18 \pm 1.69$  & $65.08 \pm 0.35$  & $4.05 \pm 0.12$  & $2.76 \pm 0.38$ & $62.91 \pm 1.24$  & $4.08 \pm 0.54$  & $4.47 \pm 0.62$ \\
      \textbf{APPNP}    & $76.79 \pm 1.84$  & $12.50 \pm 0.49$  & $16.60 \pm 2.34$  & $65.24 \pm 1.26$  & $4.52 \pm 1.02$  & $1.78 \pm 0.34$ & $67.45 \pm 1.18$  & $2.15 \pm 0.23$  & $4.35 \pm 0.76$ \\
			\textbf{FairGNN}  & $69.12 \pm 0.31$ & $12.40 \pm 0.70$  & $15.60 \pm 1.00$   & $64.04 \pm 0.90$  & $4.95 \pm 0.21$  & $4.29 \pm 0.20$  & $60.29 \pm 0.64$   & $5.30 \pm 0.20$  & $1.67 \pm 0.17$\\
      \textbf{NIFTY}    & $70.76 \pm 1.27$ & $23.26 \pm 1.35$  & $24.85 \pm 1.00$   & $65.34 \pm 0.43$  & $2.34 \pm 0.26$  & $1.46 \pm 0.27$  & $61.12 \pm 0.07$   & $6.55 \pm 0.55$  & $1.83 \pm 0.07$ \\
			\textbf{EDITS}    & $68.26 \pm 3.17$ & $21.92 \pm 0.29$  & $21.81 \pm 0.01$   & $61.60 \pm 0.54$  & $1.29 \pm 0.10$  & $1.62 \pm 0.20$  & $56.80 \pm 0.65$   & $2.75 \pm 0.80$  & $2.24 \pm 0.90$  \\
			\textbf{Graphair} & $71.49 \pm 1.00$   & $10.68 \pm 1.56$  & $12.72 \pm 2.08$   & $64.17 \pm 0.08$  & $2.10 \pm 0.17$  & $2.76 \pm 0.19$  & $62.43 \pm 0.25$   & $2.02 \pm 0.40$  & $1.62 \pm 0.47$ \\
			\textbf{BIND}     & $71.69 \pm 1.89$   & $14.37 \pm 2.62$  & $16.79 \pm 3.14$   & $63.50 \pm 0.20$  & $6.75 \pm 0.40$  & $5.41 \pm 0.57$  & $60.60 \pm 0.15$   & $5.85 \pm 0.39$  & $1.15 \pm 0.44$ \\
		  \textbf{FUGNN}   & $\mathbf{80.18 \pm 1.09}$ & $\mathbf{1.43 \pm 0.88}$ & $\mathbf{1.78 \pm 1.14}$ & $\mathbf{68.38 \pm 0.43}$ & $\mathbf{0.53 \pm 0.27}$ & $\mathbf{1.32 \pm 0.95}$ & $\mathbf{68.48 \pm 0.07}$ & $\mathbf{0.80 \pm 0.31}$ & $\mathbf{1.03 \pm 0.59}$\\
			\midrule
      \midrule
      \multirow{2}[2]{*}{\textbf{Methods}} & \multicolumn{3}{c}{\textbf{German}} & \multicolumn{3}{c}{\textbf{Bail}} & \multicolumn{3}{c}{\textbf{Credit}} \\
            \cline{2-10}
			& ACC(\%) $\uparrow$  & $\Delta_\text{SP}$(\%) $\downarrow$   & $\Delta_\text{EO}$(\%) $\downarrow$  & ACC(\%) $\uparrow$  & $\Delta_\text{SP}$(\%) $\downarrow$   & $\Delta_\text{EO}$(\%) $\downarrow$ & ACC(\%) $\uparrow$  & $\Delta_\text{SP}$(\%) $\downarrow$ & $\Delta_\text{EO}$(\%) $\downarrow$\\
			\midrule
			\textbf{GCN}      & $71.20 \pm 2.54$  & $8.43 \pm 0.44$  & $4.52 \pm 0.78$  & $89.80 \pm 1.12$  & $7.47 \pm 1.74$  & $5.23 \pm 0.78$  & $73.87 \pm 2.48$   & $12.86 \pm 1.84$  & $10.63 \pm 0.24$  \\
      \textbf{GCNII}    & $70.43 \pm 1.64$  & $2.78 \pm 0.54$  & $2.52 \pm 0.87$  & $92.43 \pm 2.37$  & $5.67 \pm 0.89$  & $3.94 \pm 0.27$  & $74.03 \pm 1.97$  & $16.85 \pm 2.54$  & $10.58 \pm 1.68$ \\
      \textbf{APPNP}    & $69.60 \pm 0.40$  & $5.29 \pm 0.16$  & $5.47 \pm 0.54$ & $85.36 \pm 2.07$  & $\mathbf{4.20 \pm 0.37}$  & $3.36 \pm 1.08$ & $74.19 \pm 0.79$  & $13.3 \pm 0.94$  & $9.47 \pm 1.86$ \\
			\textbf{FairGNN}  & $69.72 \pm 1.24$ & $3.49 \pm 0.30$  & $3.40 \pm 2.15$   & $89.68 \pm 2.09$  & $7.31 \pm 1.12$  & $5.17 \pm 0.54$  & $68.29 \pm 2.25$   & $9.74 \pm 0.28$  & $8.83 \pm 0.46$\\
      \textbf{NIFTY}    & $69.86 \pm 2.40$ & $5.73 \pm 0.55$  & $5.08 \pm 2.29$   & $81.48 \pm 2.14$  & $10.04 \pm 0.62$  & $7.71 \pm 0.10$  & $66.80 \pm 1.07$   & $13.59 \pm 0.43$  & $13.79 \pm 0.73$ \\
			\textbf{EDITS}    & $70.60 \pm 0.89$ & $4.05 \pm 1.48$  & $3.89 \pm 0.23$   & $89.57 \pm 0.46$  & $5.02 \pm 0.81$  & $2.89 \pm 0.27$  & $69.60 \pm 0.56$   & $9.13 \pm 1.38$  & $7.88 \pm 1.90$  \\
			\textbf{Graphair} & $70.08 \pm 0.64$   & $4.38 \pm 0.26$  & $3.82 \pm 0.21$   & $84.76 \pm 2.08$  & $4.98 \pm 0.34$  & $2.58 \pm 0.34$  & $67.65 \pm 2.01$   & $8.99 \pm 2.02$  & $7.05 \pm 1.74$ \\
			\textbf{BIND}     & $71.59 \pm 2.03$   & $3.46 \pm 0.10$  & $6.51 \pm 1.28$   & $88.47 \pm 2.10$  & $6.75 \pm 0.54$  & $4.23 \pm 0.32$  & $68.60 \pm 0.53$   & $11.65 \pm 0.93$  & $10.61 \pm 2.01$ \\
		  \textbf{FUGNN}   & $\mathbf{72.80 \pm 2.00}$ & $\mathbf{1.05 \pm 0.11}$ & $\mathbf{0.84 \pm 0.23}$ & $\mathbf{96.78 \pm 1.10}$ & $5.99 \pm 0.29$ & $\mathbf{1.55 \pm 0.87}$ & $\mathbf{77.02 \pm 0.07}$ & $\mathbf{0.62 \pm 0.48}$ & $\mathbf{0.18 \pm 0.09}$\\
			\bottomrule
		\end{tabular}
\label{tab:result}
\end{table*}

\begin{itemize}
  \item \textbf{Income}: Extracted from the Adult Data Set \cite{asuncion2007uci}. 
  Each node represents an individual, with connections established based on criteria similar to \cite{agarwal2021towards}. 
  The sensitive feature in this dataset is race, and the task involves classifying whether an individual's salary exceeds $50,000$ annually.
  \item \textbf{Pokec-z} and \textbf{Pokec-n}: Both datasets are collected from \textbf{Pokec}, a popular social network in Slovakia \cite{takac2012data}.
  In both datasets, each user is a node, and each edge stands for the friendship relation between two users. 
  The locating region of users is the sensitive feature.
  The task is to classify the working field of users.
  \item \textbf{Bail}: This dataset represents defendants who were released on bail at the U.S. state courts during 1990-2009 \cite{jordan2015effect}.
  Nodes represent clients within a bail bank, features are gender, loan amount, and other account-related details. 
  Edges connect clients whose credit accounts share similarities. 
  The objective is to categorize clients' credit risk as either high or low, with ``gender'' designated as the sensitive feature.
  \item \textbf{German}: Extracted from the Adult Data Set \cite{asuncion2007uci}. 
  Nodes represent defendants released in Germany from 1990 to 2009, connected by edges based on shared past criminal records and demographics. 
  The goal is predicting a defendant's likelihood of committing either a violent or nonviolent crime post-release, with ``race'' as the sensitive feature.
  \item \textbf{Credit}: Extracted from the Adult Data Set \cite{asuncion2007uci}. 
  Comprising individuals connected based on similarities in spending and payment patterns. 
  Age serves as the sensitive feature, while the label feature denotes defaulting on credit card payments.
\end{itemize}
The statistics of these six datasets are shown in Table \ref{tab:datasets}.
For datasets containing more than two classes of ground truth labels for the node classification task, we retain the class labeled $0$ and $1$, and set any class of labeled more than $1$ to $1$.
We randomly select $25$\% of nodes as the validation set and $25$\% as the test set, ensuring that the proportion of nodes labeled with each category is balanced in these sets. 
Additionally, we randomly select either $50$\% of nodes or $500$ nodes in each class of ground truth labels as the training set, depending on which is a smaller.
This partitioning strategy is consistent with prior studies \cite{agarwal2021towards, dong2022edits, dong2023interpreting}, which also serves as our baselines.
\begin{table}[H]
  \vspace{-1em}
  \centering
  \caption{The statistics of the six real-world datasets.}
  \vspace{-1em}
  \small    
  \begin{tabular}{lcccc}
    \toprule
    \textbf{Dataset} & \textbf{\# Nodes}  & \textbf{\# Edges} & \textbf{Sensitive Feature}  & \textbf{Label} \\
    \midrule
    \textbf{Income}   & $14,821$  & $51,386$  & Race    & Income  \\
    \textbf{Pokec-z}  & $67,797$  & $882,765$ & Region  & Field   \\
    \textbf{Pokec-n}  & $66,569$  & $729,129$ & Region  & Field   \\
    \textbf{German}   & $1,000$   & $24,970$  & Region  & German  \\
    \textbf{Bail}     & $18,876$  & $403,977$ & Race    & Bail    \\
    \textbf{Credit}   & $30,000$  & $200,526$ & Age     & Credit  \\
    \bottomrule
  \end{tabular}
  \vspace{-2em}
\label{tab:datasets}
\end{table}

\begin{figure*}[htbp]
	\vspace{-1em}
    \centering
	\subfigure{
		\begin{minipage}[b]{0.3 \textwidth}
			\centering
			\includegraphics[width=1\textwidth]{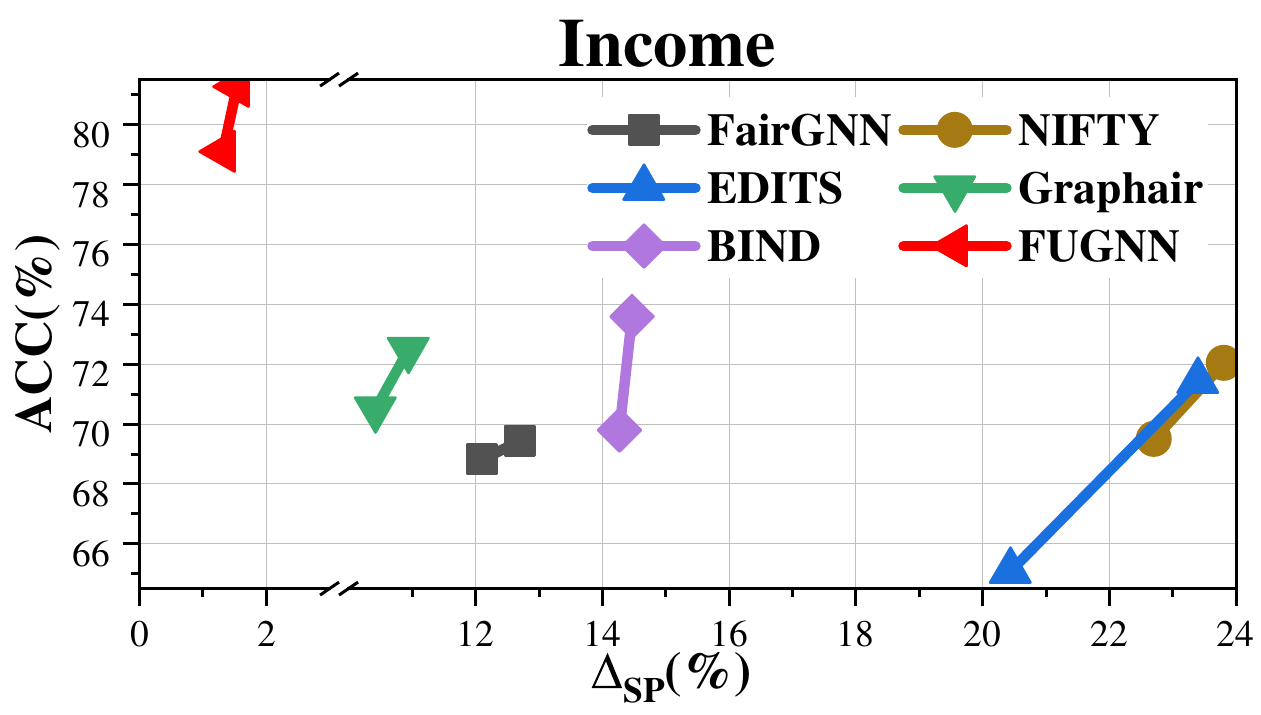}
    	\end{minipage}
	}
	\subfigure{
		\begin{minipage}[b]{0.3\textwidth}
			\centering
			\includegraphics[width=1\textwidth]{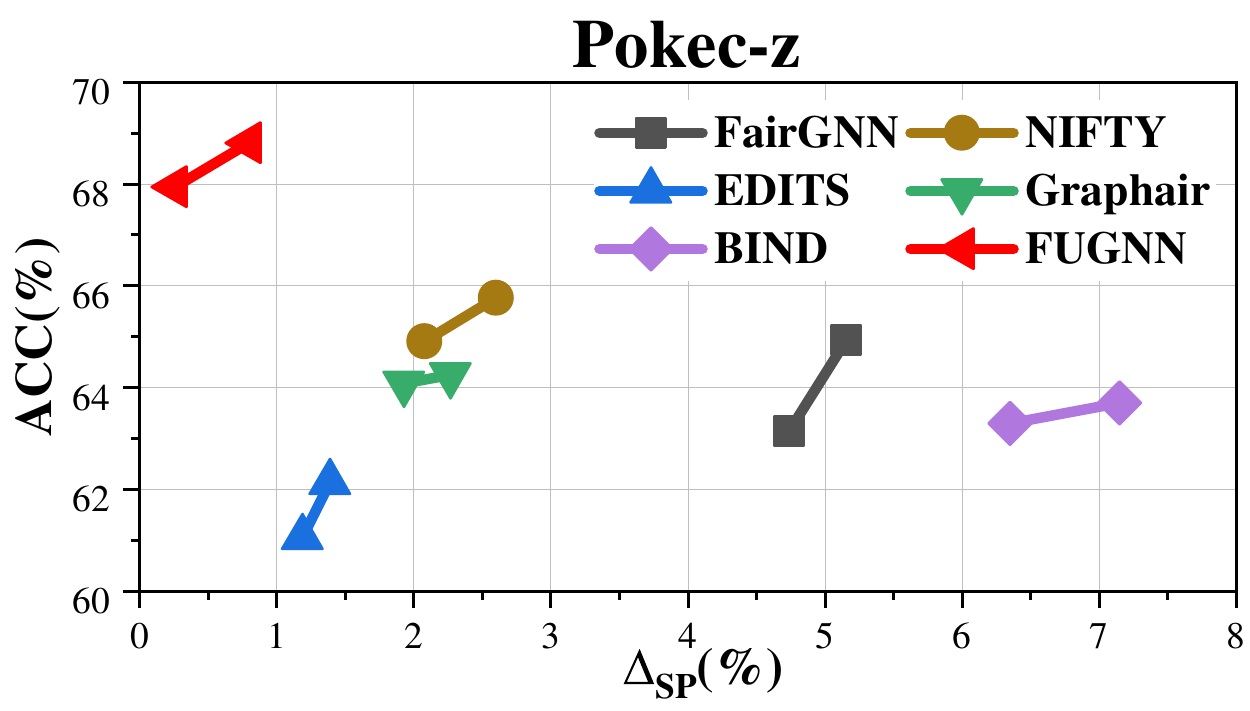}
		\end{minipage}
	}
	\subfigure{
        \begin{minipage}[b]{0.3\textwidth}
    	   \centering
    	   \includegraphics[width=1\textwidth]{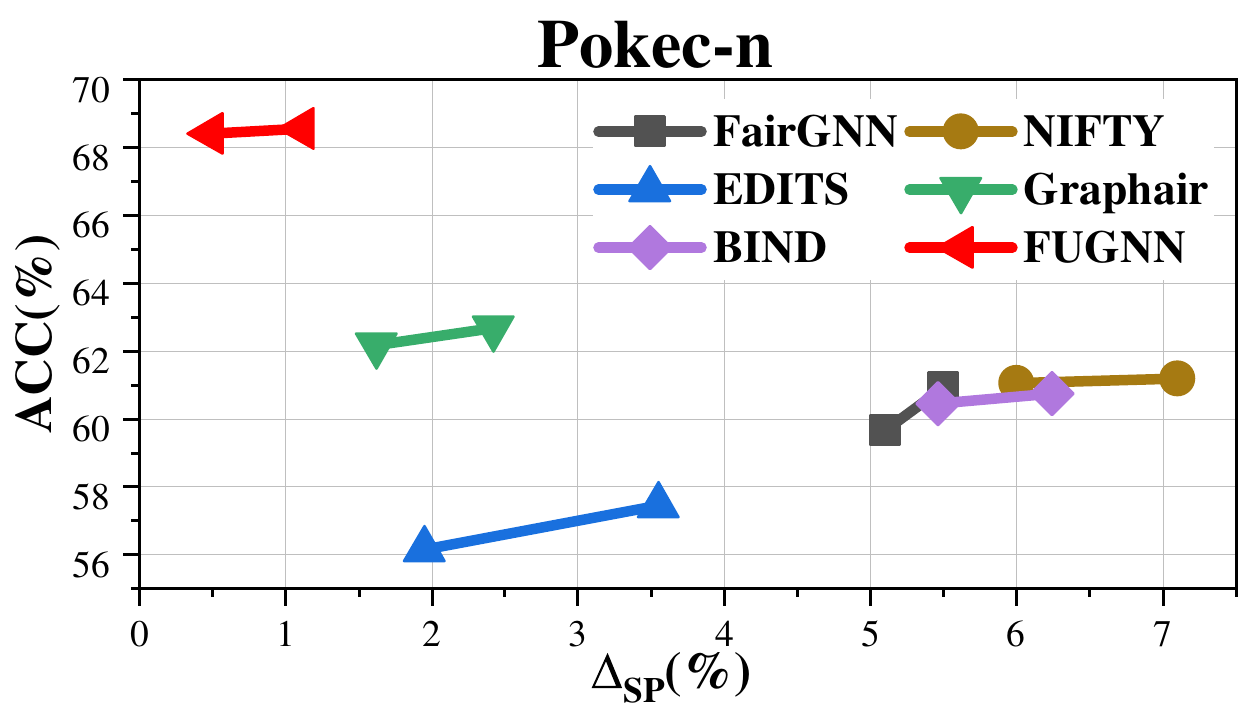}
        \end{minipage}
    }\\
    \vspace{-1.5em}
    \subfigure{
	  \begin{minipage}[b]{0.3\textwidth}
			\centering
			\includegraphics[width=1\textwidth]{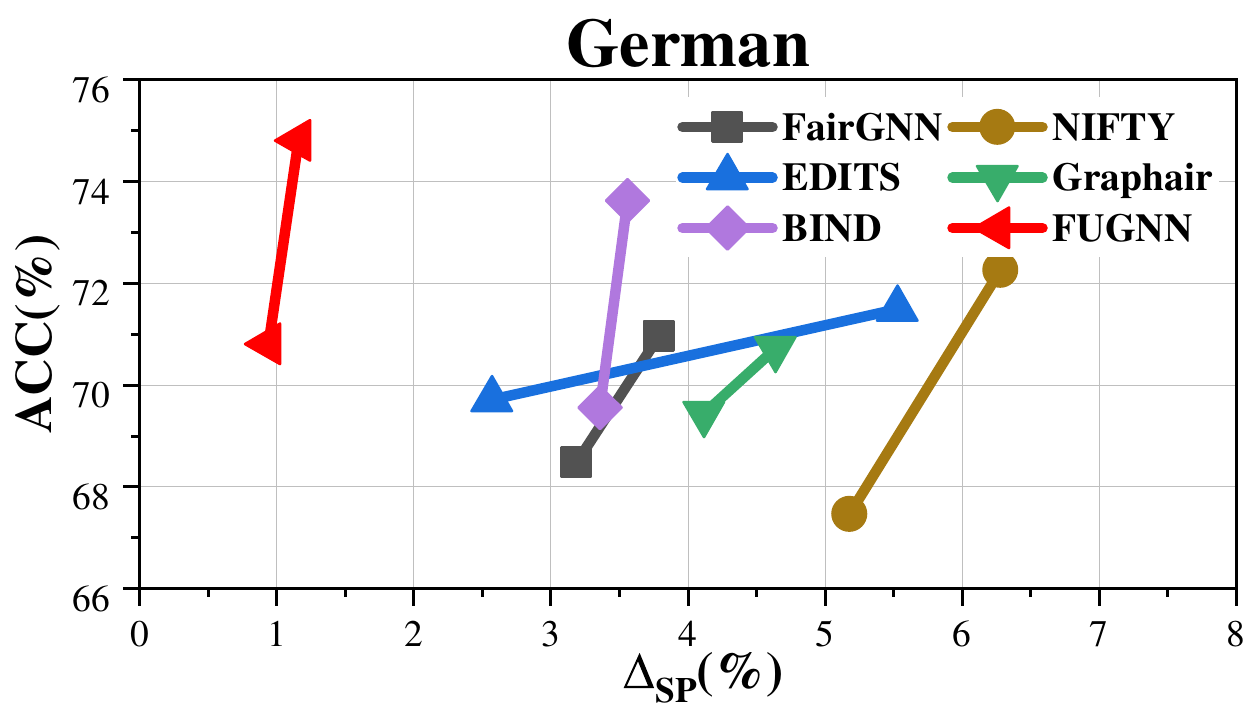}
		\end{minipage}
	}
	\subfigure{
		\begin{minipage}[b]{0.3\textwidth}
			\centering
			\includegraphics[width=1\textwidth]{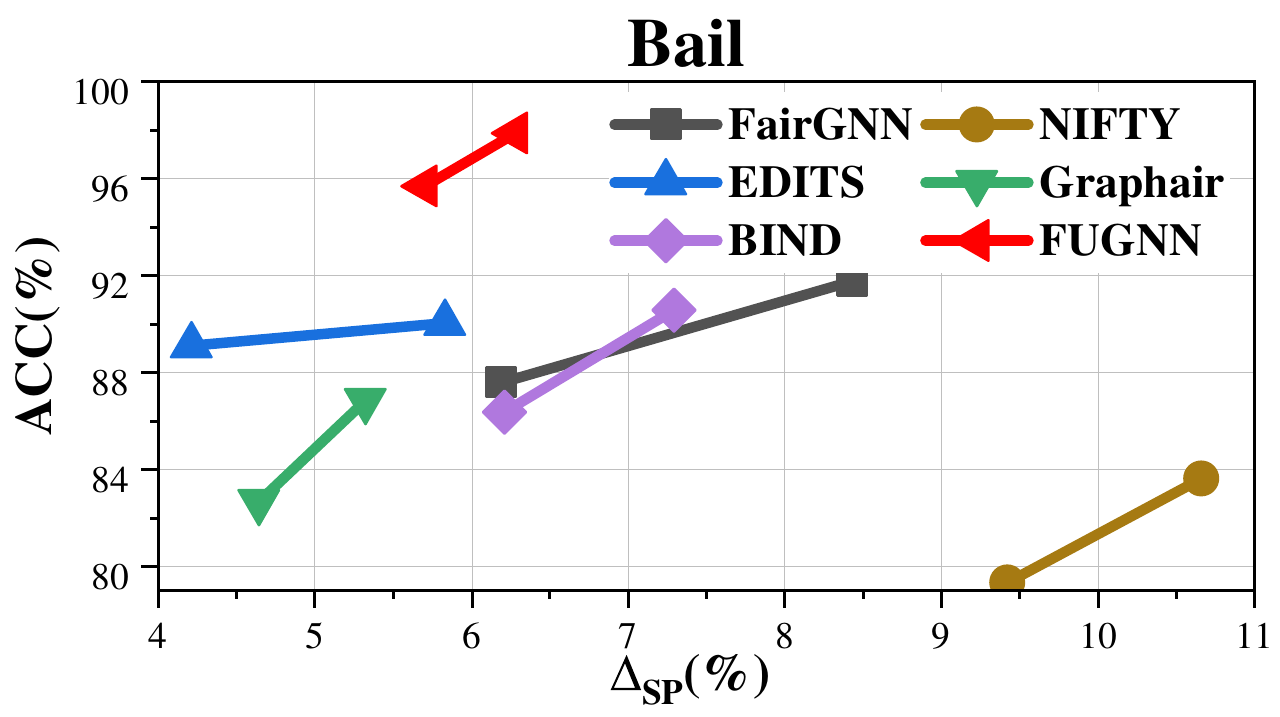}
		\end{minipage}
	}
	\subfigure{
        \begin{minipage}[b]{0.3\textwidth}
            \centering
            \includegraphics[width=1\textwidth]{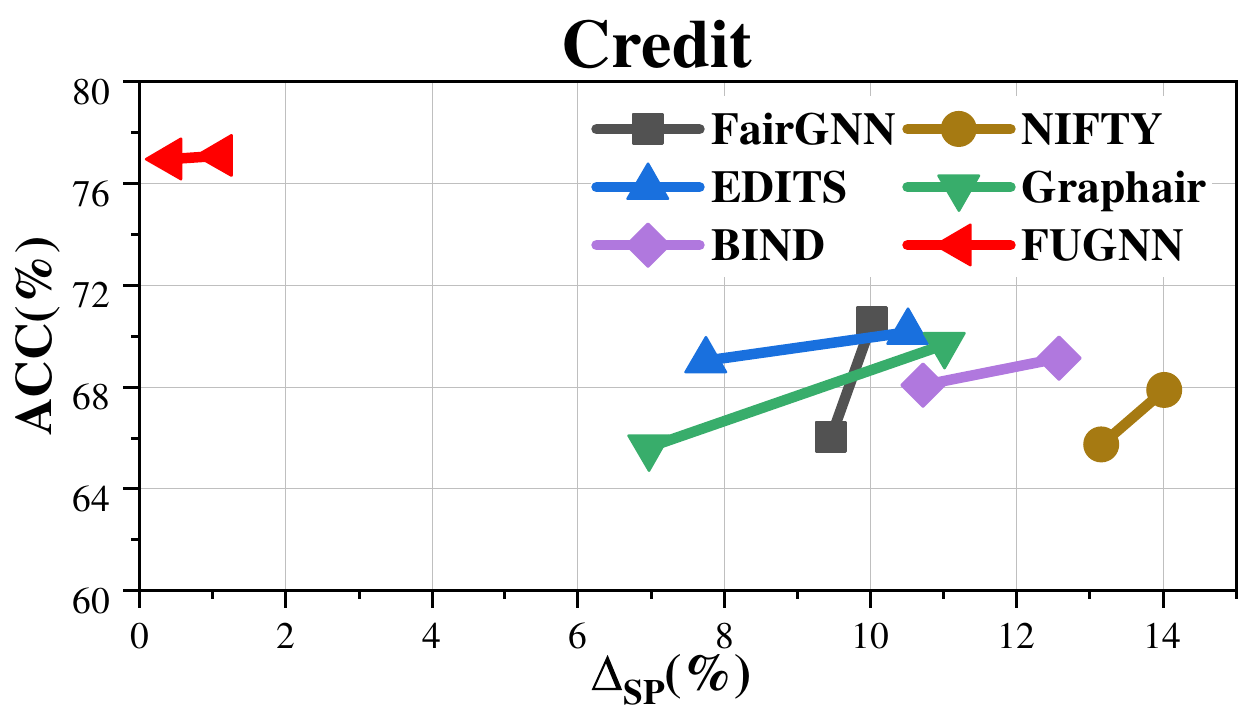}
        \end{minipage}
    }
  \vspace{-1.5em}
  \caption{The accuracy and $\Delta_\text{SP}$ trade-off. Upper-left corner is preferable.}
  \vspace{-1em}
  \label{fig:trade-off_SP}
  \vspace{-0.5em}
\end{figure*}
\begin{figure*}[htbp]
	\centering
    \subfigure{
        \begin{minipage}[b]{0.3\textwidth}
            \centering
            \includegraphics[width=1\textwidth]{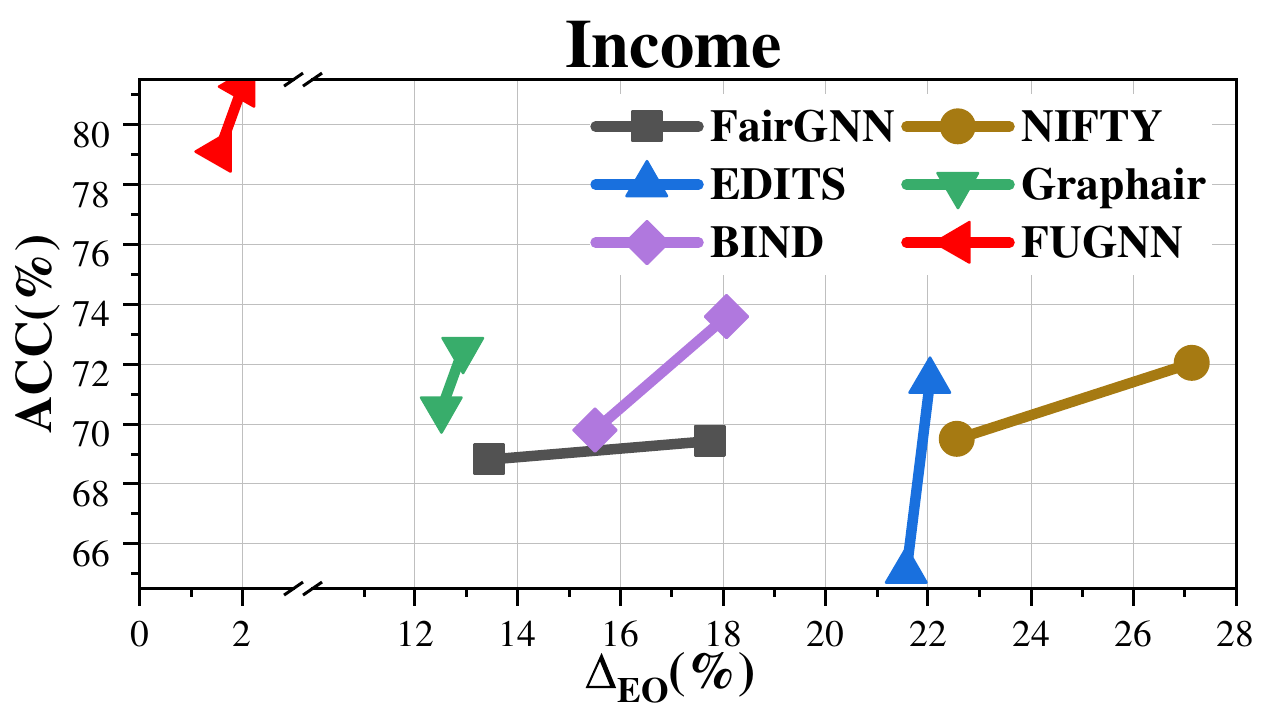}
        \end{minipage}
	}
	\subfigure{
		\begin{minipage}[b]{0.3\textwidth}
			\centering
			\includegraphics[width=1\textwidth]{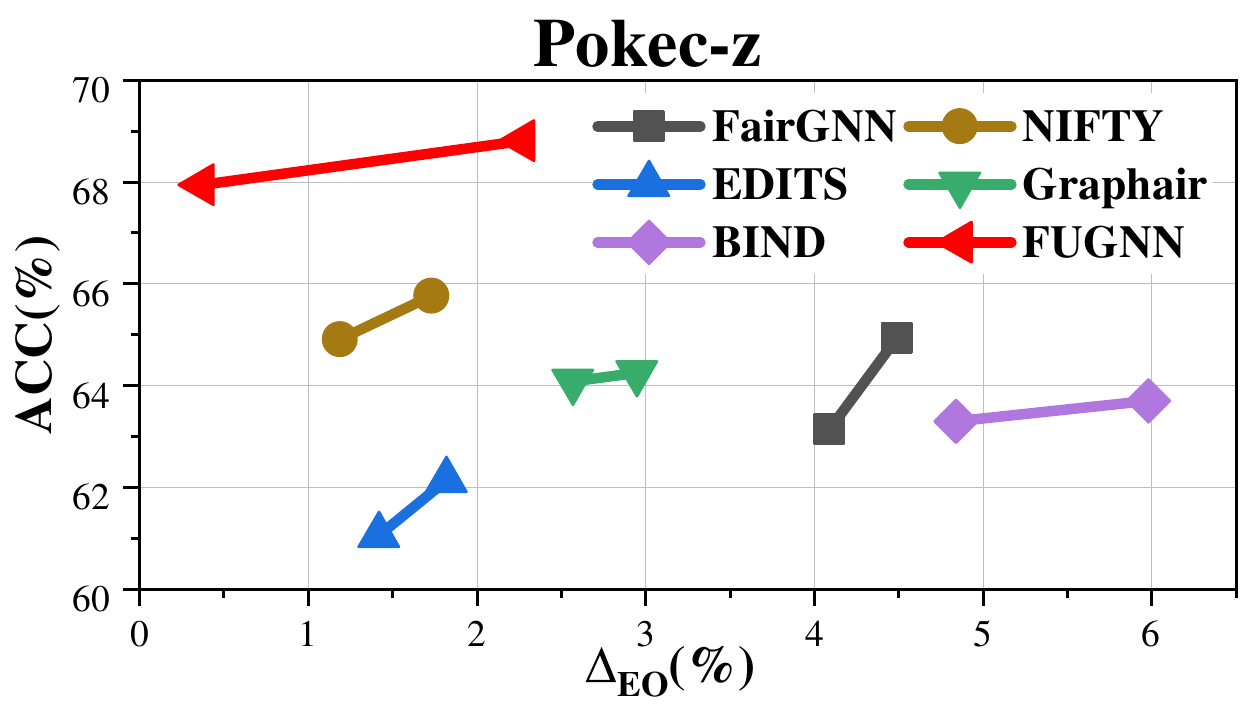}
		\end{minipage}
	}
	\subfigure{
    	\begin{minipage}[b]{0.3\textwidth}
    		\centering
    		\includegraphics[width=1\textwidth]{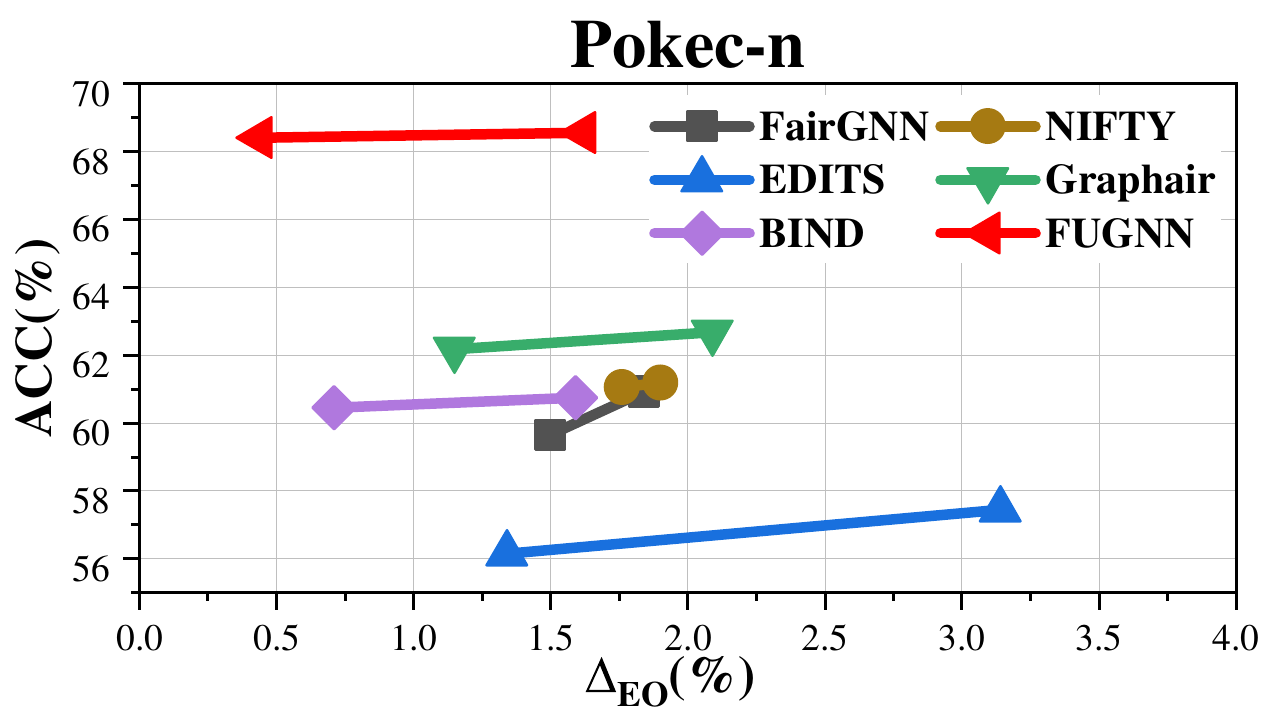}
    	\end{minipage}
    }\\
    \vspace{-1.5em}
    \subfigure{
	  \begin{minipage}[b]{0.3\textwidth}
			\centering
			\includegraphics[width=1\textwidth]{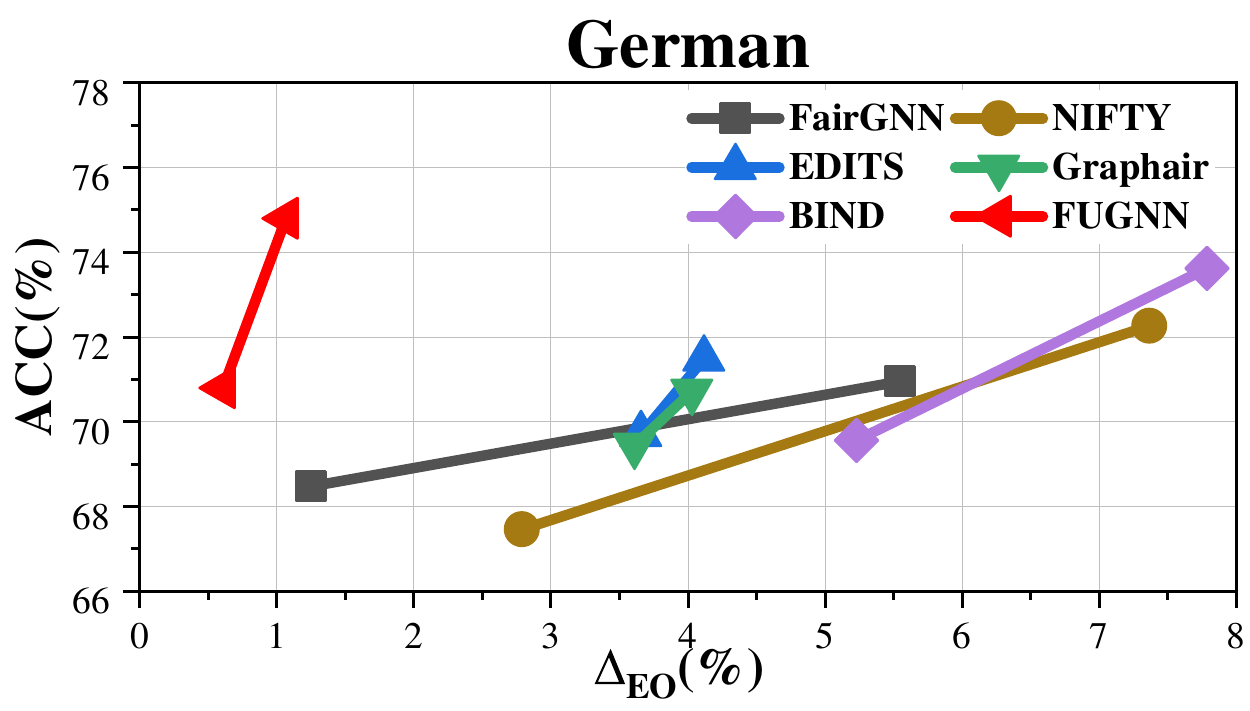}
		\end{minipage}
	}
	\subfigure{
		\begin{minipage}[b]{0.3\textwidth}
			\centering
			\includegraphics[width=1\textwidth]{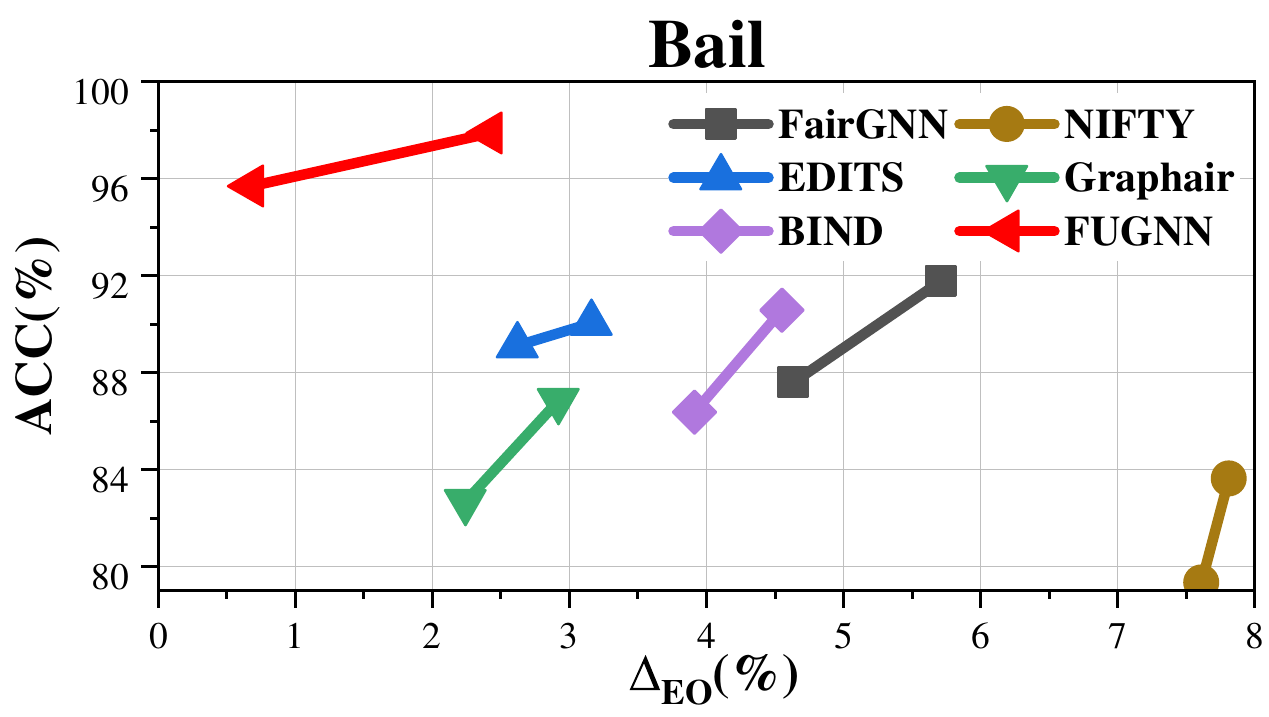}
		\end{minipage}
	}
	\subfigure{
    \begin{minipage}[b]{0.3\textwidth}
      \centering
      \includegraphics[width=1\textwidth]{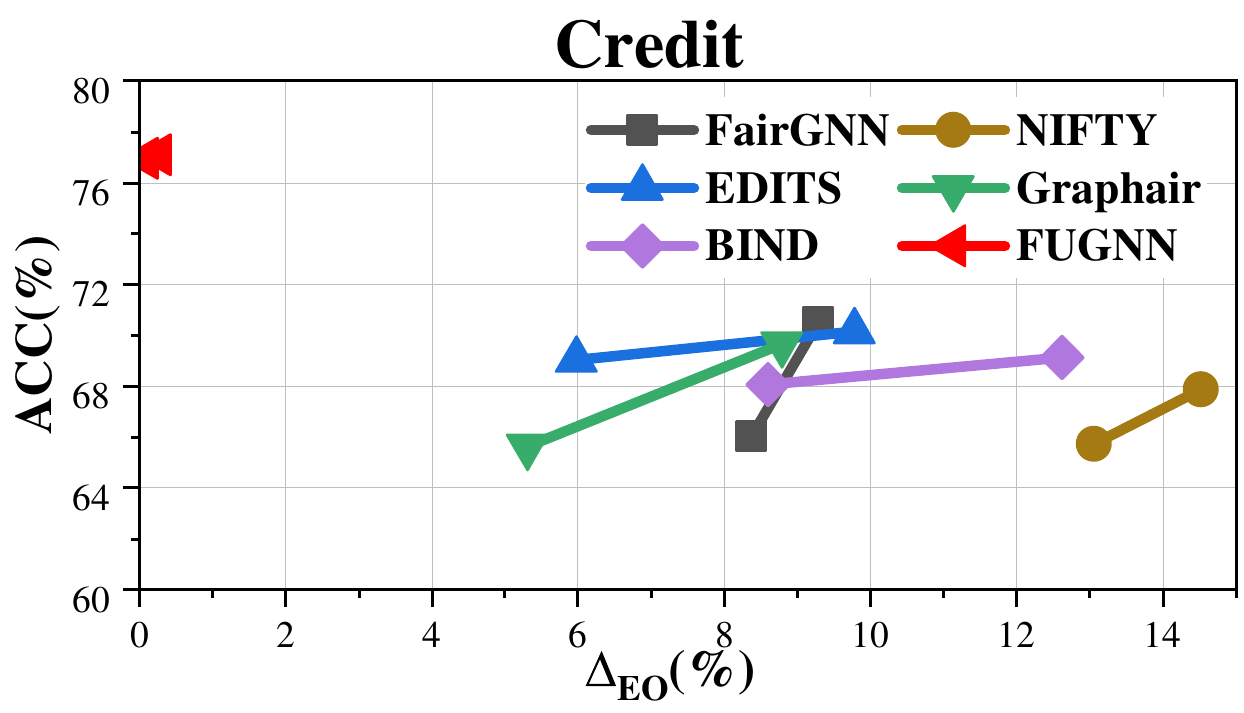}
    \end{minipage}
    }
    \vspace{-1.5em}
  \caption{The accuracy and $\Delta_\text{EO}$ trade-off. Upper-left corner is preferable.}
  \vspace{-1em}
  \label{fig:trade-off_EO}
\end{figure*}
\subsection{Baselines}
We compare our proposed method with three spectral convolutional network methods: \textbf{GCN} \cite{kipf2017semi}, \textbf{GCNII} \cite{chen2020simple}, and \textbf{APPNP} \cite{klicpera2019predict}, as well as the following representative and state-of-the-art fairness-aware GNNs methods:
\begin{itemize}
    \item \textbf{NIFTY} \cite{agarwal2021towards} seeks to optimize the alignment between predictions derived from perturbed sensitive features and those generated using unperturbed features.
    \item \textbf{EDITS} \cite{dong2022edits} involves pre-processing the input graph data to reduce bias by employing feature and structural debiasing methods.
    \item \textbf{FairGNN} \cite{dai2023learning} utilizes adversarial training to eliminate sensitive feature information from node embeddings.
    \item \textbf{Graphair} \cite{ling2023learning} focuses on acquiring equitable utility through automated graph data augmentations.
    \item \textbf{BIND} \cite{dong2023interpreting} effectively estimates the impact of each training node on the disparity in probabilistic distributions. 
\end{itemize}

The optimal hyperparameters for all methods are obtained by grid search.
We employ the released implementations of fairness-aware GNN baselines, namely FairGNN, NIFTY, EDITS, BIND, and Graphair, to ensure a fair and consistent comparison. 
All baselines are implemented using the PyTorch framework \cite{paszke2019pytorch}.
Specifically, FairGNN, NIFTY, BIND and Graphair are optimized with Adam optimizer \cite{kingma2015adam}, while EDITS utilizes RMSprop \cite{hinton2012neural} as recommended.
For each method, we conduct experiments with different seeds \{$0$, $1$, $2$, $3$, $4$\} and use the mean value and standardized covariance as the results.
All models are implemented using PyTorch and PyTorch-geometric \cite{fey2019fast}, executed on the Ubuntu 20.04.6 LTS operation system, with hardware specifications including an Intel(R) Xeon(R) Silver 4114 CPU @ 2.20GHz and a Tesla V100S-PCIE-32G GPU.

\subsection{Comparison Results}
For assessing utility, we adopt node classification accuracy as the corresponding metric, while for assessing fairness, we adopt two traditional metrics $\Delta_\text{SP}$ and $\Delta_\text{EO}$.
Higher accuracy indicates better utility, and lower values of fairness metrics indicate better fairness.

\subsubsection{Fairness}
Table \ref{tab:result} provides a comprehensive overview of the fairness evaluation metrics for our proposed FUGNN method and various baseline models across six real-world datasets.
FUGNN consistently demonstrates outstanding fairness utility, particularly notable in terms of $\Delta_\text{SP}$ and $\Delta_\text{EO}$ across the evaluated datasets. 
An exception is observed in the $\Delta_\text{SP}$ for the \textbf{Bail} dataset, which shows poor performance. 
This can be partly attributed to excessive accuracy of the model. 
However, the lower $\Delta_\text{EO}$ indicates that the algorithm ensures fairness in predicting the correct samples, reaffirming its effectiveness in addressing fairness-related concerns within graph-based learning algorithms.

\subsubsection{Trade-off between accuracy and fairness}
To confirm our pursuit of a comprehensive assessment in accuracy and fairness, we aim to achieve the best fairness with the highest accuracy.
From Table \ref{tab:result}, we observe that our proposed FUGNN improves both fairness and utility simultaneously.
The utility achieved is even higher than the three spectral graph convolution network methods.
To provide a more visual representation of our effectiveness, we present the details in Figure \ref{fig:trade-off_SP} and Figure \ref{fig:trade-off_EO}. 
For the fairness evaluation, we employ the $\Delta_\text{SP}$ and $\Delta_\text{EO}$ metrics, respectively.
Notably, the upper-left corner point within these graphical utilities symbolizes the optimal utility, characterized by the highest accuracy and the highest prediction fairness.
This achievement aligns with the primary objective of our approach, which aims to concurrently enhance both the fairness and utility of GNNs.
\begin{table*}[htbp]
  \centering
  \small
	\caption{Comparisons among different components in the FUGNN model. OOM denotes out-of-memory.}
  \vspace{-1em}
  \tabcolsep=0.2cm
	\begin{tabular}{cccccccccccccc}
    \toprule
			\multirow{2}[2]{*}{\textbf{Datasets}} & \multicolumn{3}{c}{\textbf{FUGNN}} & \multicolumn{3}{c}{\textbf{FUGNN w/o FES}} & \multicolumn{3}{c}{\textbf{FUGNN w/o OED}} \\
    \cline{2-10}
      & ACC(\%) $\uparrow$  & $\Delta_\text{SP}$(\%) $\downarrow$   & $\Delta_\text{EO}$(\%) $\downarrow$  & ACC(\%) $\uparrow$  & $\Delta_\text{SP}$(\%) $\downarrow$   & $\Delta_\text{EO}$(\%) $\downarrow$ & ACC(\%) $\uparrow$  & $\Delta_\text{SP}$(\%) $\downarrow$ & $\Delta_\text{EO}$(\%) $\downarrow$\\
    \midrule
			\textbf{Income}   & $\mathbf{80.18 \pm 1.09}$ & $\mathbf{1.43 \pm 0.88}$ & $\mathbf{1.78 \pm 1.14}$ & $76.75 \pm 1.87$ & $3.79 \pm 0.96$ & $5.78 \pm 1.90$ & $79.98 \pm 0.60$ & $2.29 \pm 0.34$ & $3.26 \pm 0.59$ \\
      \textbf{Pokec-z}  & $\mathbf{68.38 \pm 0.43}$ & $\mathbf{0.53 \pm 0.27}$ & $\mathbf{1.32 \pm 0.95}$ & OOM & OOM & OOM & $67.26 \pm 0.33$ & $1.04 \pm 0.37$ & $3.43 \pm 0.42$\\
      \textbf{Pokec-n}  & $\mathbf{68.48 \pm 0.07}$ & $\mathbf{0.80 \pm 0.31}$ & $\mathbf{1.03 \pm 0.59}$ & OOM & OOM & OOM & $67.97 \pm 0.27$ & $2.37 \pm 0.39$ & $2.97 \pm 1.16$\\
      \textbf{German}   & $\mathbf{72.80 \pm 2.00}$ & $\mathbf{1.05 \pm 0.11}$ & $\mathbf{0.84 \pm 0.23}$ & $69.20 \pm 0.40$ & $5.07 \pm 1.23$ & $4.83 \pm 0.94$ & $68.90 \pm 1.15$ & $2.20 \pm 0.62$ & $4.60 \pm 1.13$\\
      \textbf{Bail}  & $\mathbf{96.78 \pm 1.10}$ & $\mathbf{5.99 \pm 0.29}$ & $\mathbf{1.55 \pm 0.87}$ & $96.39 \pm 0.14$ & $6.51 \pm 0.16$ & $2.35 \pm 0.53$ & $96.59 \pm 0.27$ & $6.53 \pm 0.23$ & $1.86 \pm 0.23$\\
      \textbf{Credit}  & $\mathbf{77.02 \pm 0.07}$ & $\mathbf{0.62 \pm 0.48}$ & $\mathbf{0.18 \pm 0.09}$ & $74.39 \pm 1.58$ & $3.05 \pm 0.36$ & $3.35 \pm 0.74$ & $75.91 \pm 1.11$ & $2.82 \pm 0.40$ & $1.99 \pm 0.61$\\
    \bottomrule
  \end{tabular}
  \label{tab:ablation}
\end{table*}
\vspace{-1em}
\subsection{Ablation Study}
FUGNN, as an approach to harmonize the fairness and utility in GNNs, achieves its objectives through two key components: $\text{FUGNN}_\text{FES}$ and $\text{FUGNN}_\text{OED}$. 
To comprehensively assess the contributions of these two elements and the overall impact of FUGNN, we conduct an ablation study. 
This study aims to discern whether both $\text{FUGNN}_\text{FES}$ and $\text{FUGNN}_\text{OED}$ contribute to enhancing prediction fairness and accuracy.
In the ablation study, we systematically remove each component independently to assess their individual impacts. 

We first focus on the contribution of eigenvalue calculation, represented as \textbf{FUGNN w/o FES} (FUGNN without $\text{FUGNN}_\text{FES}$). 
Notably, when comparing $\text{FUGNN}_\text{FES}$ with FUGNN, the latter consistently outperforms them in terms of statistical parity, equal opportunity, and accuracy.
The results, shown in Table \ref{tab:ablation}, reinforce the necessity of $\text{FUGNN}_\text{FES}$ within FUGNN.
We then denote the configuration without optimization of eigenvector distribution as \textbf{FUGNN w/o OED} (FUGNN without $\text{FUGNN}_\text{OED}$). 
The results, as presented in Table \ref{tab:ablation}, provide compelling evidence of the contributions of this component.
When comparing FUGNN to the ablated versions, FUGNN is consistently higher in terms of accuracy and lower in terms of $\Delta_\text{SP}$ and $\Delta_\text{EO}$. 
This observation underscores that $\text{FUGNN}_\text{OED}$ plays a crucial role in enhancing both the fairness of GNN predictions and accuracy.

In summary, the ablation study collectively emphasizes the integral role played by $\text{FUGNN}_\text{FES}$ and $\text{FUGNN}_\text{OED}$ in improving fairness and utility in GNNs.
\vspace{-1em}
\subsection{Parameter Analysis}
\label{sec:PA}
In this investigation, our primary aim is to scrutinize the impact of parameter $K$ on our proposed FUGNN. 
Specifically, we conduct a systematic parameter study that focuses on the variable denoted as $K$ across the six datasets.
The parameter $K$ assumes a pivotal role as it governs how original information from the adjacency matrix is retained.
Moreover, it also plays a crucial role in representing the similarity between original sensitive features and ones after $l$ layers of convolution.
According to our theoretical analysis, as $K$ increases, fairness initially hovers, followed by a subsequent decline.
The following experiment results verify our analysis.
\begin{table}[H]
  \centering
  \small
  \tabcolsep=0.1cm
  \vspace{-2em}
  \caption{The accuracy, $\Delta_\text{SP}$ and $\Delta_\text{EO}$ of FUGNN w.r.t. different parameter $k$ values.}
  \vspace{-1em}
  \begin{tabular}{lcccccc}
    \toprule
    \multirow{2}{*}{$\mathbf{K}$} & \multicolumn{3}{c}{\textbf{Income}} & \multicolumn{3}{c}{\textbf{Credit}} \\
    \cline{2-7}
    & ACC(\%) $\uparrow$  & $\Delta_\text{SP}$(\%) $\downarrow$   & $\Delta_\text{EO}$(\%) $\downarrow$  & ACC(\%)$\uparrow$  & $\Delta_\text{SP}$(\%) $\downarrow$   & $\Delta_\text{EO}$(\%) $\downarrow$\\
    \midrule
    \textbf{1}      & $78.68$ & $0.01$ & $1.66$ & $77.04$ & $0.04$ & $0.15$\\
    \textbf{2}      & $78.95$ & $0.51$ & $0.73$ & $76.99$ & $0.57$ & $0.13$  \\
    \textbf{3}      & $78.33$ & $1.98$ & $0.79$ & $77.14$ & $0.64$ & $0.72$\\
    \textbf{4}      & $80.19$ & $1.33$ & $3.24$ & $77.11$ & $0.66$ & $0.12$  \\
    \textbf{5}      & $80.46$ & $1.16$ & $1.14$ & $76.95$ & $1.20$ & $0.32$\\
    \textbf{6}      & $78.33$ & $2.27$ & $0.03$ & $77.04$ & $0.04$ & $0.15$  \\
    \textbf{7}      & $81.06$ & $2.04$ & $2.25$ & $76.99$ & $0.48$ & $0.01$\\
    \textbf{8}      & $80.90$ & $1.70$ & $1.81$ & $76.89$ & $1.08$ & $0.61$  \\
    \textbf{9}      & $79.57$ & $0.91$ & $0.25$ & $76.92$ & $0.35$ & $0.29$\\
    \textbf{10}     & $78.44$ & $0.02$ & $1.06$ & $76.64$ & $0.89$ & $0.37$  \\
    \midrule
    \textbf{100}    & $77.20$ & $2.38$ & $3.36$ & $74.89$ & $2.09$ & $2.38$\\
    \textbf{500}    & $77.33$ & $2.56$ & $3.36$ & $74.15$ & $2.61$ & $2.11$  \\
    \textbf{1,000}  & $77.22$ & $4.22$ & $2.40$ & $73.14$ & $2.07$ & $2.46$  \\
    \textbf{5,000}  & $77.14$ & $5.47$ & $4.48$ & $72.81$ & $2.69$ & $2.81$\\
    \textbf{$n$}    & $76.75$ & $3.79$ & $5.78$ & $74.39$ & $3.05$ & $3.35$  \\
    \bottomrule
  \end{tabular}
  \vspace{-1em}
  \label{tab:parameter analysis}
\end{table}
\vspace{-1.5em}

\begin{table}[H]
  \centering
  \small
  \tabcolsep=0.1cm
  \vspace{-1.5em}
  \caption{The accuracy, $\Delta_\text{SP}$ and $\Delta_\text{EO}$ of FUGNN w.r.t. different parameter $k$ values. OOM denotes out-of-memory.}
  \vspace{-1em}
  \begin{tabular}{lcccccc}
    \toprule
    \multirow{2}{*}{$\mathbf{K}$} & \multicolumn{3}{c}{\textbf{Pokec-z}} & \multicolumn{3}{c}{\textbf{Pokec-n}} \\
    \cline{2-7}
    & ACC(\%) $\uparrow$  & $\Delta_\text{SP}$(\%) $\downarrow$   & $\Delta_\text{EO}$(\%) $\downarrow$  & ACC(\%)$\uparrow$  & $\Delta_\text{SP}$(\%) $\downarrow$   & $\Delta_\text{EO}$(\%) $\downarrow$\\
    \midrule
    \textbf{1}      & $67.54$ & $0.70$ & $2.62$ & $68.59$ & $0.24$ & $0.12$\\
    \textbf{2}      & $67.61$ & $1.20$ & $0.89$ & $67.32$ & $0.68$ & $0.38$\\
    \textbf{3}      & $68.12$ & $1.32$ & $0.26$ & $67.91$ & $1.05$ & $0.31$\\
    \textbf{4}      & $68.08$ & $1.87$ & $0.19$ & $68.00$ & $0.66$ & $0.62$\\
    \textbf{5}      & $68.55$ & $2.51$ & $0.04$ & $68.05$ & $1.18$ & $0.99$\\
    \textbf{6}      & $68.20$ & $0.99$ & $0.04$ & $68.68$ & $1.98$ & $1.43$\\
    \textbf{7}      & $68.36$ & $1.60$ & $0.15$ & $68.27$ & $3.18$ & $4.32$\\
    \textbf{8}      & $67.46$ & $2.38$ & $0.41$ & $68.72$ & $2.31$ & $1.70$\\
    \textbf{9}      & $68.51$ & $0.83$ & $1.29$ & $68.45$ & $1.78$ & $2.46$\\
    \textbf{10}     & $67.73$ & $1.23$ & $0.35$ & $68.36$ & $3.68$ & $5.40$\\
    \midrule
    \textbf{100}    & $66.79$ & $3.67$ & $4.03$ & $67.10$ & $4.17$ & $5.79$\\
    \textbf{500}    & $66.43$ & $4.26$ & $4.51$ & $66.78$ & $4.31$ & $5.90$\\
    \textbf{1,000}  & $65.98$ & $4.34$ & $4.51$ & $66.43$ & $4.21$ & $6.55$\\
    \textbf{5,000}  & $65.56$ & $4.74$ & $5.02$ & $65.97$ & $5.73$ & $6.78$\\
    \textbf{$n$}    & OOM     & OOM    & OOM    & OOM     & OOM    & OOM   \\
    \bottomrule
  \end{tabular}
  \vspace{-2em}
  \label{tab:parameter analysis 2}
\end{table}

\begin{table}[H]
  \centering
  \small
  \tabcolsep=0.1cm
  \vspace{-1em}
  \caption{The accuracy, $\Delta_\text{SP}$ and $\Delta_\text{EO}$ of FUGNN w.r.t. different parameter $k$ values.}
  \vspace{-1em}
  \begin{tabular}{lcccccc}
    \toprule
    \multirow{2}{*}{$\mathbf{K}$} & \multicolumn{3}{c}{\textbf{German}} & \multicolumn{3}{c}{\textbf{Bail}} \\
    \cline{2-7}
    & ACC(\%) $\uparrow$  & $\Delta_\text{SP}$(\%) $\downarrow$   & $\Delta_\text{EO}$(\%) $\downarrow$  & ACC(\%)$\uparrow$  & $\Delta_\text{SP}$(\%) $\downarrow$   & $\Delta_\text{EO}$(\%) $\downarrow$\\
    \midrule
    \textbf{1}      & $71.20$ & $1.94$ & $3.36$ & $97.32$ & $6.15$ & $1.41$\\
    \textbf{2}      & $71.20$ & $0.23$ & $3.36$ & $97.64$ & $6.15$ & $1.60$\\
    \textbf{3}      & $72.80$ & $1.05$ & $0.84$ & $98.23$ & $6.13$ & $1.35$\\
    \textbf{4}      & $71.60$ & $0.85$ & $0.84$ & $97.65$ & $6.10$ & $1.73$\\
    \textbf{5}      & $71.60$ & $1.27$ & $0.84$ & $97.18$ & $6.25$ & $1.50$\\
    \textbf{6}      & $68.80$ & $0.67$ & $5.78$ & $97.77$ & $5.73$ & $0.88$\\
    \textbf{7}      & $70.80$ & $0.85$ & $1.68$ & $97.35$ & $6.26$ & $1.56$\\
    \textbf{8}      & $70.00$ & $0.42$ & $0.84$ & $97.94$ & $6.26$ & $1.49$\\
    \textbf{9}      & $70.00$ & $0.85$ & $1.68$ & $97.25$ & $6.47$ & $1.91$\\
    \textbf{10}     & $71.60$ & $0.45$ & $1.58$ & $98.11$ & $6.26$ & $1.81$\\
    \midrule
    \textbf{100}    & $70.80$ & $3.01$ & $6.72$ & $97.89$ & $6.22$ & $2.07$\\
    \textbf{500}    & $69.20$ & $3.45$ & $5.67$ & $96.89$ & $6.30$ & $2.09$\\
    \textbf{1,000}  & $69.20$ & $5.07$ & $4.83$ & $96.21$ & $6.24$ & $2.45$\\
    \textbf{5,000}  & -       & -      & -      & $96.91$ & $6.48$ & $2.51$\\
    \textbf{$n$}    & $69.20$ & $5.07$ & $4.83$ & $96.39$ & $6.51$ & $2.35$\\
    \bottomrule
  \end{tabular}
  \vspace{-1em}
  \label{tab:parameter analysis 3}
\end{table}
Our analysis involves exploring $K$ values spanning the range of 1-10, 100, 500, 1,000, 5,000, and $n$. 
To show how $K$ affects $\Delta_\text{SP}$ and $\Delta_\text{EO}$, we present the average results for multiple runs.
The outcomes of \textbf{Income} and \textbf{Credit} are shown in Table \ref{tab:parameter analysis}.
Additional results of the parameter analysis are shown in Table \ref{tab:parameter analysis 2} and Table \ref{tab:parameter analysis 3}.
Given that the number of nodes in the \textbf{German} dataset is $1,000$, far below $5,000$, there is no $K=5,000$ result available for the \textbf{German} dataset, and the $K=1,000$ result is congruent with the $K=n$ result.
Additionally, we conduct eigedecomposition on the adjacency matrices of \textbf{Pokec-z} and \textbf{Pokec-n}, resulting in processed eigenvalue and eigenvector files of approximately 18 GB in size. 
When we applied these files for model training, we encountered Out-Of-Memory (OOM) errors.
The findings indicate that as $K$ varies in $\{1,2,...,10\}$, both $\Delta_\text{SP}$ and $\Delta_\text{EO}$ exhibit relatively consistent performance with moderate fluctuations. When $K$ reaches >100, the model performance shows a clear declination in accuracy and fairness.
This experiment observation aligns with the analysis of $K$ in \textit{Lemma 3}.
After comparing both $\Delta_\text{SP}$ and $\Delta_\text{EO}$, we choose the optimal result as our final selection for the parameter $K$.
Specifically, we select $K=1$ for \textbf{Income}, $K=3$ for \textbf{Pokec-z}, $K=2$ for \textbf{Pokec-n}, $K=10$ for \textbf{German}, $K=3$ for \textbf{Bail}, and $K=6$ for \textbf{Credit}.

It worths mentioning that the best number of principal eigenvectors to be kept varies across datasets. 
Hence, conducting comprehensive hyperparameter search would be beneficial to run FUGNN effectively on new datasets.
This also suggests a potential direction for further improving FUGNN via developing efficient strategies for parameter refining.

\subsection{Training Cost Comparison}
The whole eigendecomposition causes cubic complexity in terms of the number of nodes, resulting in a computational cost of $O(n^3)$.
The eigenvector selection eliminates the need to rank the eigenvectors following the entire eigendecomposition process.
The Arnolodi Package algorithm, which $\text{FUGNN}_\text{FES}$ employs, achieves a time complexity of $O(nK^2)$ for eigendecomposition.
In particular, FUGNN only chooses principal eigenvectors, ensuring that $K$ remains consistently below $10$.
We present the comparison results of whole eigendecomposition (WE) and $\text{FUGNN}_\text{FES}$ is shown in Table \ref{tab:runtime eig}.
\begin{table}[H]
  \vspace{-1em}
  \small
  \caption{Runtime (s) of WE and $\text{FUGNN}_\text{FES}$.}
  \vspace{-1em}
  \centering
  \tabcolsep=0.1cm
  \begin{tabular}{lcccccc}
    \toprule
    \textbf{Methods} & \textbf{Income} & \textbf{Pokec-z} & \textbf{Pokec-n} & \textbf{German} & \textbf{Bail} & \textbf{Credit} \\
    \midrule
    \textbf{WE}   & $107.81$   & $3018.91$   & $2987.16$   & $0.46$ & $198.89$ & $774.91$\\
    \textbf{$\text{FUGNN}_\text{FES}$}  & $0.84$ & $10.00$ & $8.84$ & $0.36$ & $3.29$ & $0.84$  \\
    \bottomrule
  \end{tabular}
  \vspace{-1em}
  \label{tab:runtime eig}
\end{table}

According to the results, we can observe that the time savings are more significant with larger datasets. 
Especially for datasets where using the full decomposition results in inference exceeding the available memory limits, our method still works fine in such cases.

\section{Conclusion}
In this work, we examine the protection of sensitive features via studying the similarity between the representations of those sensitive features at the input layer and at deep latent space of the model after $l$ layers of convolution operations. We establish a strong correlation with the largest magnitude eigenvalue of the adjacency matrix. 
Drawing inspiration from this finding, we have presented FUGNN, an innovative approach dedicated to enhancing both the fairness and utility of GNNs. 
FUGNN is built on two crucial components: eigenvalue adjustment and optimization of eigenvector distribution. 
These two components are designed based on our theoretical findings, which simultaneously enhance the fairness and utility of GNNs. The proposed FUGNN framework has demonstrated significant improvements over strong baselines across diverse real-world scenarios. 
In future work, we plan to further refine the efficiency of the FUGNN approach and extend its applicability to situations with limited sensitive features.

\bibliographystyle{ACM-Reference-Format}
\bibliography{arXiv_FUGNN}
\end{document}